\documentclass[journal]{IEEEtran}

\ifCLASSINFOpdf
\else
\fi

\usepackage{framed,multirow}
\usepackage{graphicx}
\usepackage{amssymb}
\usepackage{amsmath}
\usepackage{multirow}
\usepackage{booktabs}
\usepackage{array}
\usepackage{todonotes}
\newcolumntype{L}[1]{>{\raggedright\let\newline\\\arraybackslash\hspace{0pt}}m{#1}}
\newcolumntype{C}[1]{>{\centering\let\newline\\\arraybackslash\hspace{0pt}}m{#1}}
\newcolumntype{R}[1]{>{\raggedleft\let\newline\\\arraybackslash\hspace{0pt}}m{#1}}

\usepackage{lipsum}
\usepackage{changepage}
\usepackage{color,soul}
\usepackage{amssymb}
\usepackage{amsmath}
\usepackage{enumitem}   

\usepackage{float}

\usepackage{booktabs,xcolor}
\definecolor{lightgray}{gray}{0.9}

\usepackage{tikz}

\usepackage{lipsum}

\let\OLDthebibliography\thebibliography
\renewcommand\thebibliography[1]{
  \OLDthebibliography{#1}
  \setlength{\parskip}{0pt}
  \setlength{\itemsep}{0pt plus 0.3ex}
}

\usepackage{array}

\makeatletter
\newcommand{\thickhline}{%
    \noalign {\ifnum 0=`}\fi \hrule height 1pt
    \futurelet \reserved@a \@xhline
}
\newcolumntype{"}{@{\hskip\tabcolsep\vrule width 1pt\hskip\tabcolsep}}
\makeatother

\usepackage{amssymb}
\usepackage{latexsym}

\usepackage{url}
\usepackage{xcolor}

\usepackage{hyperref}

\definecolor{newcolor}{rgb}{.8,.349,.1}

\begin{document}

\onecolumn
\begin{center}
\Huge{Deep learning with noisy labels: \\ exploring techniques and remedies in medical image analysis}
\end{center}

\vspace{1mm}

\begin{center}
\normalsize
Davood Karimi, Haoran Dou, Simon K. Warfield, and Ali Gholipour \\ Department of Radiology, Boston Children’s Hospital, Harvard Medical School, Boston, MA, USA
\end{center}

\vspace{2mm}

\begin{abstract}

Supervised training of deep learning models requires large labeled datasets. There is a growing interest in obtaining such datasets for medical image analysis applications. However, the impact of label noise has not received sufficient attention. Recent studies have shown that label noise can significantly impact the performance of deep learning models in many machine learning and computer vision applications. This is especially concerning for medical applications, where datasets are typically small, labeling requires domain expertise and suffers from high inter- and intra-observer variability, and erroneous predictions may influence decisions that directly impact human health. In this paper, we first review the state-of-the-art in handling label noise in deep learning. Then, we review studies that have dealt with label noise in deep learning for medical image analysis. Our review shows that recent progress on handling label noise in deep learning has gone largely unnoticed by the medical image analysis community. To help achieve a better understanding of the extent of the problem and its potential remedies, we conducted experiments with three medical imaging datasets with different types of label noise, where we investigated several existing strategies and developed new methods to combat the negative effect of label noise. Based on the results of these experiments and our review of the literature, we have made recommendations on methods that can be used to alleviate the effects of different types of label noise on deep models trained for medical image analysis. We hope that this article helps the medical image analysis researchers and developers in choosing and devising new techniques that effectively handle label noise in deep learning.

\end{abstract}

\footnotesize
\hspace{5mm} \textbf{Index Terms: } label noise  deep learning  machine learning  big data  medical image annotation




\vspace{5mm}

\begin{figure*}[hbt!]
\begin{minipage}[b]{1.0\linewidth}
  \centering
  \centerline{\includegraphics[width=16.0cm]{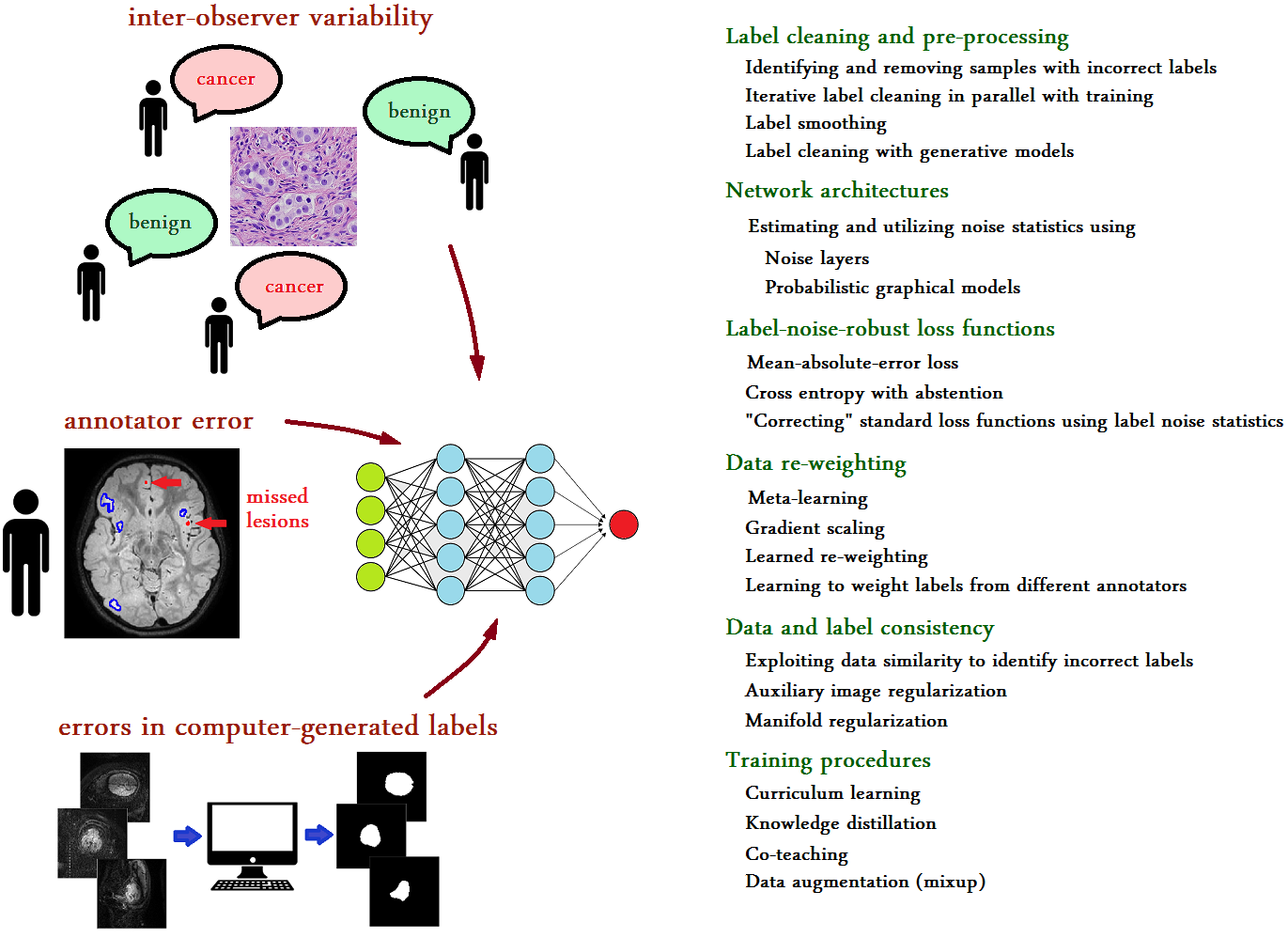}}
  \end{minipage}
\caption{Label noise is a common feature of medical image datasets. Left: The major sources of label noise include inter-observer variability, human annotator's error, and errors in computer-generated labels. The significance of label noise in such datasets is likely to increase as larger datasets are prepared for deep learning. Right: A quick overview of possible strategies to deal with, or to account for label noise in deep learning.}
\label{fig:methods_schematics}
\end{figure*}

\normalsize
\twocolumn
\section{Introduction}

\subsection{Background}

Deep learning has already made an impact on many branches of medicine, in particular medical imaging, and its impact is only expected to grow \cite{ching2018,topol2019}. Even though it was first greeted with much skepticism  \cite{wang2017}, in a few short years it proved itself to be a worthy player in solving many problems in medicine, including problems in disease and patient classification, patient treatment recommendation, outcome prediction, and more \cite{ching2018}. Many experts believe that deep learning will play an important role in the future of medicine and will be an enabling tool in medical research and practice \cite{topol2019b, prevedello2019}. With regard to medical image analysis, methods that use deep learning have already achieved impressive, and often unprecedented, performance in many tasks ranging from low-level image processing tasks such as denoising, enhancement, and reconstruction \cite{wang2018d}, to more high-level image analysis tasks such as segmentation, detection, classification, and registration \cite{ronneberger2015,haskins2019}, and even more challenging tasks such as discovering links between the content of medical images and patient's health and survival \cite{xu2019, mobadersany2018}.

The recent success of deep learning has been attributed to three main factors \cite{lecun2015,sun2017}. First, technical advancements in network architecture design, network parameter initialization, and training methods. Second, increasing availability of more powerful computational hardware, in particular graphical processing units and parallel processing, that allow training of very large models on massive datasets. Last, but not least, increasing availability of very large and growing datasets. However, even though in some applications it has become possible to curate large datasets with reliable labels, in most applications it is very difficult to collect and accurately label datasets large enough to effortlessly train deep learning models. A solution that is becoming more popular is to employ non-expert humans or automated systems with little or no human supervision to label massive datasets \cite{guo2016,deng2009,ipeirotis2010}. However, datasets collected using such methods typically suffer from very high label noise \cite{wang2018c, kuznetsova2018}, thus they have limited applicability in medical imaging.

The challenge of obtaining large datasets with accurate labels is particularly significant in medical imaging. The available data is typically small to begin with, and data access is hampered by such factors as patient privacy and institutional policies. Furthermore, labeling of medical images is very resource-intensive because it depends on domain experts. In some applications, there is also significant inter-observer variability among experts, which will necessitate obtaining consensus labels or labels from multiple experts and proper methods of aggregating those labels \cite{bridge2016,nir2018}. Some studies have been able to employ a large number of experts to annotate large medical image datasets \cite{gulshan2016,esteva2017}. However, such efforts depend on massive financial and logistical resources that are not easy to obtain in many domains. Alternatively, a few studies have successfully used automated mining of medical image databases such as hospital picture archiving and communication systems (PACS) to build large training datasets \cite{yan2018,irvin2019}. However, this method is not always applicable as historical data may not include all the desired labels or images. Moreover, label noise in such datasets is expected to be higher than in expert-labeled datasets. There have also been studies that have used crowd-sourcing methods to obtain labels from non-experts \cite{gurari2015, albarqouni2016}. Even though this method may have potential for some applications, it has a limited scope because in most medical applications non-experts are unable to provide useful labels. Even for relatively simple segmentation tasks, computerized systems have been shown to generate significantly less accurate labels compared with human experts and crowdsourced non-experts \cite{gurari2015}. In general, lack of large datasets with trustworthy labels is considered to be one of the biggest challenges facing a wider adoption and successful deployment of deep learning methods in medical applications \cite{langlotz2019, ching2018, ravi2016}.

\subsection{Aims and scope of this paper}

Given the outline presented above, it is clear that relatively small datasets with noisy labels are, and will continue to be, a common scenario in training deep learning models in medical image analysis applications. Hence, algorithmic approaches that can effectively handle the label noise are highly desired. In this manuscript, we first review and explain the recent advancements in training deep learning models in the presence of label noise. We review the methods proposed in the general machine learning literature, most of which have not yet been widely employed in medical imaging applications. Then, we review studies that have addressed label noise in deep learning with medical imaging data. Finally, we present the results of our experiments on three medical image datasets with noisy labels, where we investigate the performance of several strategies to deal with label noise, including a number of new methods that we have developed for each application. Based on our results, we make general recommendations to improve deep learning with noisy training labels in medical imaging data.

In the field of medical image analysis, in particular, the notion of label noise is elusive and not easy to define. The term has been used in the literature to refer to different forms of label imperfections or corruptions. Especially in the era of big data, label noise may manifest itself in various forms. Therefore, at the outset we need to clarify the intended meaning of label noise in this paper and demarcate the scope of this study to the extent possible. 

To begin with, it should be clear that we are only interested in label noise, and not data/measurement noise. Specifically, consider a set $\{x_i, y_i \}$ of medical images, $x_i$, and their corresponding labels, $y_i$. Although $x_i$ may include measurement noise, that is not the focus of this review. We are only interested in the noise in the label, $y_i$. Typically, the label $y$ is a discrete variable and can be either an image-wise label, such as in classification problems, or a pixel/voxel-wise label, such as in dense segmentation. Moreover, in this paper we are only concerned with labeled data. Semi-supervised methods are methods that use both labeled and unlabeled training data. Many semi-supervised methods synthesize (noisy) labels for unlabeled data, which are then used for training. Such studies fall within the scope of this study if they use novel or sophisticated methods to handle noisy synthesized labels. Another form of label imperfection that is becoming more common in medical image datasets is when there is only image-level label, and no pixel-level annotations are available \cite{wang2017a, irvin2019}. This type of label is referred to as weak label and is used by methods that are termed weakly supervised learning or multiple-instance learning methods. This type of label imperfection is also beyond the scope of this study. Luckily, there are recent review articles that cover these types of label imperfections. Semi-supervised learning, multiple-instance learning, and transfer learning in medical image analysis have been reviewed in \cite{cheplygina2019}. Focusing only on medical image segmentation, another recent paper reviewed methods for dealing with scarce and imperfect annotations in general, including weak and sparse annotations \cite{tajbakhsh2019}.

The organization of this article is as follows. In Section \ref{label_noise_in_ML} we briefly describe methods for handling label noise in classical (i.e., pre-deep learning) machine learning. In Section \ref{label_noise_in_DL} we review studies that have dealt with label noise in deep learning. Then, in Section \ref{label_noise_in_MIA} we take a closer look into studies that have trained deep learning models on medical image datasets with noisy labels. Section \ref{experiments} contains our experimental results with three medical image datasets, where we investigate the impact of label noise and the potential of techniques and remedies for dealing with noisy labels in deep learning. Conclusions are presented in Section \ref{conclusions}.

\section{Label noise in classical machine learning}
\label{label_noise_in_ML}

Learning from noisy labels has been a long-standing challenge in machine learning \cite{frenay2013,garcia2015}. Studies have shown that the negative impact of label noise on the performance of machine learning methods can be more significant than that of measurement/feature noise \cite{zhu2004b,quinlan1986}. The complexity of label noise distribution varies greatly depending on the application. In general, label noise can be of three different types: class-independent (the simplest case), class-dependent, and class and feature-dependent (potentially much more complicated). Most of the methods that have been proposed to handle noisy labels in classical machine learning fall into one of the following three categories \cite{frenay2013}:

\begin{enumerate}

\item Methods that focus on model selection or design. Fundamentally, these methods aim at selecting or devising models that are more robust to label noise. This may include selecting the model, the loss function, and the training procedures. It has been known that the impact of label noise depends on the type and design of the classifier model. For example, naive Bayes and random forests are more robust than other common classifiers such as decision trees and support vector machines \cite{nettleton2010,folleco2008}, and that boosting can exacerbate the impact of label noise \cite{abellan2010,mcdonald2003,long2010b}, whereas bagging is a better way of building classifier ensembles in the presence of significant label noise \cite{dietterich2000}. Studies have also shown that 0-1 label loss is more robust than smooth alternatives (e.g., exponential loss, log-loss, squared loss, and hinge-loss) \mbox{\cite{manwani2013,patrini2016}}. Other studies have modified standard loss functions to improve their robustness to label noise, for example by making the hinge loss negatively unbounded as proposed in \cite{van2015b}. Furthermore, it has been shown that proper re-weighting of training samples can improve the robustness of many loss functions to label noise \cite{liu2015,natarajan2013}.

\item Methods that aim at reducing the label noise in the training data. A popular approach is to train a classifier using the available training data with noisy labels or a small dataset with clean labels and identify mislabeled data samples based on the predictions of this classifier \cite{segata2009}. Voting among an ensemble of classifiers has been shown to be an effective method for this purpose \cite{brodley1996,sluban2010}. K-nearest neighbors (KNN)-based analysis of the training data has also been used to remove mislabeled instances \cite{wilson1997,wilson2000}. More computationally intensive approaches include those that identify mislabeled instances via their impact on the training process. For example, \cite{zhang2009,malossini2006} propose to detect mislabeled instances based on their impact on the classification of other instances in a leave-one-out framework. Some methods are similar to outlier-detection techniques. They define some criterion to reflect the classification uncertainty or complexity of a data point and prune those training instances that exceed a certain threshold on that criterion \cite{gamberger2000,sun2007}.

\item Methods that perform classifier training and label noise modeling in a unified framework. Methods in this class can overlap with those of the two aformentioned classes. For instance, some methods learn to denoise labels or to identify and down-weight samples that are more likely to have incorrect labels in parallel with classifier training. Some methods in this category improve standard classifiers such as support vector machines, decision trees, and neural networks by proposing novel training procedures that are more robust to label noise \cite{khardon2007,lin2004}. Alternatively, different forms of probabilistic models have been used to model the label noise and thereby improve various classifiers \cite{kaster2010,kim2006}.

\end{enumerate}

\section{Deep learning with noisy labels}
\label{label_noise_in_DL}

Deep learning models typically require much more training data than the more traditional machine learning models do. In many applications the training data are labeled by non-experts or even by automated systems. Therefore, the label noise level is usually higher in these datasets compared with the smaller and more carefully prepared datasets used in classical machine learning.

Many recent studies have demonstrated the negative impact of label noise on the performance of deep learning models and have investigated the nature of this impact. It has been shown that, even with regularization, current convolutional neural network (CNN) architectures used for image classification and trained with standard stochastic gradient descent (SGD) algorithms can fit very large training datasets with completely random labels \cite{zhang2016}. Obviously, the test performance of such a model would be similar to random assignment because the model has only memorized the training data. Given such an enormous representation capacity, it may seem surprising that large deep learning models have achieved record-breaking performance in many real-world applications. The answer to this apparent contradiction, as suggested by \cite{arpit2017}, is that when deep learning models are trained on typical datasets with mostly correct labels, they do not memorize the data. Instead, at least in the beginning of training, they learn the dominant patterns shared among the data samples. It has been conjectured that this behavior is due to the distributed and hierarchical representation inherent in the design of the state of the art deep learning models and the explicit regularization techniques that are commonly used when training them \cite{arpit2017}. One study empirically confirmed these ideas by showing that deep CNNs are robust to strong label noise \cite{rolnick2017}. For example, in hand-written digit classification on the MNIST dataset, if the label accuracy was only $1 \%$ higher than random labels, a classification accuracy of $90 \%$ was achieved at test time. A similar behavior was observed on more challenging datasets such as CIFAR100 and ImageNet, albeit at much lower label noise levels. This suggests strong learning (as opposed to memorization) tendency of large CNNs. However, somewhat contradictory results have been reported by other studies. For face recognition, for example, it has been found that label noise can have a significant impact on the accuracy of a CNN and that training on a smaller dataset with clean labels is better than training on a much larger dataset with significant label noise \cite{wang2018c}. The theoretical reasoning and experiments in \cite{chen2019b} suggested a quadratic relation between the label noise ratio in the training data and test error. 

Although the details of the interplay between memorization and learning mentioned above is not fully understood, experiments in \cite{arpit2017} suggest that this trade-off depends on the nature and richness of the data, amount of label noise, model architecture, as well as training procedures including regularization. \cite{ma2018} show that the local intrinsic dimensionality of the features learned by a deep learning model depends on the label noise. Formal definition of local intrinsic dimensionality is given by \mbox{\cite{houle2017}}. It quantifies the dimensionality of the underlying data manifold. More specifically, given a data point $x_i$, local intrinsic dimensionality of the data manifold is a measure of the rate of encounter of other data points as the radius of a ball centered at $x_i$ grows. \cite{ma2018} showed that when training on data with noisy labels, the local dimensionality of the features initially decreases as the model learns the dominant patterns in the data. As the training proceeds, the model begins to overfit to the data samples with incorrect labels and the dimensionality starts to increase. \cite{drory2018} establish an analogy between the performance of deep learning models and KNN under label noise. Using this analogy, they empirically show that deep learning models are highly sensitive to label noise that is concentrated, but that they are less sensitive when the label noise is spread across the training data.

The theoretical work on understanding the impact of label noise on the training and generalization of deep neural networks is still ongoing \cite{martin2017}. On the practical side, many studies have shown the negative impact of noisy labels on the performance of these models in real-world applications \cite{yu2017a,moosavi2017,speth2019}. Not surprisingly, therefore, this topic has been the subject of much research in recent years. We review some of these studies below, organizing them under six categories. As this categorization is arbitrary, there is much overlap among the categories and some studies may be argued to belong to more than one category.

Table 1 shows a summary of the methods we have reviewed. For each category of methods, we have shown a set of representative studies along with the applications addressed in the experimental results of the original paper. For each category of methods, we have also suggested some applications in medical image analysis that can benefit from the methods developed in those papers.

\begin{table*}[!htb]
\footnotesize

\begin{tabular}{ L{2.1cm}  L{9.0cm} L{5.7cm} }
\thickhline
 Methods category & Representative studies from machine learning and computer vision literature & Potential applications in medical image analysis   \\ \thickhline

\multirow{5}{*}{\parbox{2cm}{Label cleaning and pre-processing}} & \cite{ostyakov2018} - image classification  & \multirow{5}{*}{\parbox{5.5cm}{most applications, including disease and pathology classification (\cite{pham2019}; experiments in Section \ref{pathology_experiment}) and lesion detection and segmentation (experiments in Section \ref{tsc_experiment})}} \\
& \cite{lee2018} - image classification  &  \\
& \cite{northcutt2017} - image classification  &  \\
& \cite{veit2017} - image classification  &  \\
& \cite{gao2017} - regression, classification, semantic segmentation & \\
\hline

\multirow{3}{*}{\parbox{2cm}{Network architecture}} & \cite{sukhbaatar2014b} -  image classification  & \multirow{3}{*}{\parbox{5.5cm}{lesion detection (\cite{dgani2018}), pathology classification (experiments in Section \ref{pathology_experiment}) }}  \\
&  \cite{vahdat2017} - image classification &  \\
&  \cite{yao2018} - image classification &  \\
\hline

\multirow{6}{*}{Loss functions} &  \cite{ghosh2017} - image and text classification &   \multirow{6}{*}{\parbox{5.5cm}{ lesion detection (experiments in Section \ref{tsc_experiment}), pathology classification (experiments in Section \ref{pathology_experiment}),  segmentation (\cite{matuszewski2018}; experiments in Section \ref{segmentation_experiment})}}  \\
& \cite{zhang2018} - image classification  &  \\
& \cite{wang2019c} - image classification, object detection  &  \\
& \cite{rusiecki2019} - image classification & \\
& \cite{boughorbel2018} - electronic health records & \\
& \cite{hendrycks2018} - image and text classification & \\
\hline

\multirow{5}{*}{\parbox{2cm}{Data re-weighting}} & \cite{ren2018} -  image classification  & \multirow{5}{*}{\parbox{5.5cm}{ lesion detection (\cite{le2019}) and segmentation (experiments in Section \ref{tsc_experiment}), lesion classification (\cite{xue2019}; experiments in Section \ref{pathology_experiment}), segmentation (\cite{zhu2019}; \cite{mirikharaji2019}) }}  \\
& \cite{shu2019} - image classification  &  \\
& \cite{khetan2017} - image classification  &  \\
& \cite{tanno2019}  - image classification &  \\
& \cite{shen2019} -  image classification & \\
\hline

\multirow{6}{*}{\parbox{2cm}{ Data and label consistency}} &  \cite{lee2019} - image classification  &  \multirow{6}{*}{\parbox{5.5cm}{lesion detection and classification, segmentation (\cite{yu2019b}) }}  \\
& \cite{zhang2019} - image classification &  \\
& \cite{speth2019} - facial attribute recognition  &  \\
& \cite{azadi2015} - image classification  &  \\
& \cite{wang2018b}- image classification  &  \\
& \cite{reed2014} - image classification, emotion recognition, object detection  &  \\
\hline

\multirow{7}{*}{\parbox{2cm}{Training procedures}} & \cite{zhong2019} - face recognition  & \multirow{7}{*}{\parbox{5.5cm}{most applications, including segmentation (experiments in Section \ref{segmentation_experiment}; \cite{min2018}; \cite{nie2018}; \cite{zhang2018b}), lesion detection (experiments in Section \ref{tsc_experiment}), and classification (\cite{fries2019})}}  \\
& \cite{jiang2017} - image classification  &  \\
& \cite{sukhbaatar2014b} - image classification  &  \\
& \cite{han2018} - image classification & \\
& \cite{zhang2017} - image classification & \\
& \cite{acuna2019} - boundary segmentation & \\
& \cite{yu2018} - boundary segmentation & \\
\thickhline
 
\end{tabular}

    \label{table:summary_of_methods}

 \caption{\small{Summary of the main categories of methods for learning with noisy labels, representative studies, and potential applications in medical image analysis. The left column indicates the six categories under which we classify the studies reviewed in Sections~\mbox{\ref{label_noise_in_ML}} and ~\mbox{\ref{label_noise_in_DL}}. The middle column lists several representative studies from the fields of machine learning and computer vision and the applications considered in those studies. The right column suggests potential applications for the methods in each category in medical image analysis. In this column, where applicable, we have cited relevant published studies from the field of medical image analysis and experiments reported in Section \mbox{\ref{experiments}} of this paper as examples of the application of methods adapted or developed in each category.}}

\end{table*}

\subsection{Label cleaning and pre-processing}

The methods in this category aim at identifying and either fixing or discarding training data samples that are likely to have incorrect labels. This can be done either prior to training or iteratively in parallel with the training of the main model. \cite{vo2015} proposed supervised and unsupervised image ranking methods for identifying correctly-labeled images in a large corpus of images with noisy labels. The proposed methods were based on matching each image with a noisy label to a set of representative images with clean labels. This method improved the classification accuracy by 4-6\% over the baseline CNN models on three datasets. \cite{veit2017} trained two CNNs in parallel using a small dataset with correct labels and a large dataset with noisy labels. The two CNNs shared the feature extraction layers. One CNN used the clean dataset to learn to clean the noisy dataset, which was used by the other CNN to learn the main classification task. Experiments showed that this training method was more effective than training on the large noisy dataset followed by fine-tuning on the clean dataset. \cite{ostyakov2018} trained an ensemble of classifiers on data with noisy labels using cross-validation and used the predictions of the ensemble as soft labels for training the final classifier.

CleanNet, proposed by \cite{lee2018}, extracts a feature vector from a query image with a noisy label and compares it with a feature vector that is representative of its class. The representative feature vector for each class is computed from a small clean dataset. The similarity between these feature vectors is used to decide whether the label is correct. Alternatively, this similarity can be used to assign weights to the training samples, which is the method proposed for image classification by \cite{lee2018}. \cite{han2019} improved upon CleanNet in several ways. Most importantly, they removed the need for a clean dataset by estimating the correct labels in an iterative framework. Moreover, they allowed for multiple prototypes (as opposed to only one in CleanNet) to represent each class. Both of these studies reported improvements in image classification accuracy of 1-5\% depending on the dataset and noise level.

A number of proposed methods for label denoising are based on classification confidence. Rank Pruning, proposed by \cite{northcutt2017}, identifies data points with confident labels and updates the classifier using only those data points. This method is based on the assumption that data samples for which the predicted probability is close to one are more likely to have correct labels. However, this is not necessarily true. In fact, there is extensive recent work showing that standard deep learning models are not ``well calibrated" \cite{guo2017,lakshminarayanan2017}. A classifier is said to have a calibrated prediction confidence if its predicted class probability indicates its likelihood of being correct. For a perfectly-calibrated classifier, $P \big( y_{\text{predicted}}=y_{\text{true}} | \hat{p}=p \big) = p$. It has been shown that deep learning models produce highly over-confident predictions. Many studies in recent years have aimed at improving the calibration of deep learning models \cite{gal2015,kendall2017,pawlowski2017}. In order to reduce the reliance on classifier calibration, the Rank Pruning algorithm, as its name suggests, ranks the data samples based on their predicted probability and removes the data samples that are least confident. In other words, Rank Pruning assumes that the predicted probabilities are accurate in the relative sense needed for ranking. In light of what is known about poor calibration of deep learning models, this might still be a strong assumption. Nonetheless, Rank Pruning was shown empirically to lead to substantial improvements in image classification tasks in the presence of strong label noise. Identification of incorrect labels based on prediction confidence was also shown to be highly effective in extensive experiments on image classification by \cite{ding2018}, improving the classification accuracy on CIFAR-10 by up to 20\% in the presence of very strong label nosie. \cite{kohler2019} proposed an iterative label noise filtering approach based on similar concepts as Rank Pruning. This method estimates prediction uncertainty (using such methods as Deep Ensembles \cite{lakshminarayanan2017} or Monte-Carlo dropout \cite{kendall2017}) during training and relabels data samples that are likely to have incorrect labels.

A different approach, is proposed by \cite{gao2017}. In this approach, termed deep label distribution learning (DLDL), the initial noisy labels are smoothed to obtain a ``label distribution", which is a discrete distribution for classification problems. The authors propose methods for obtaining this label distribution from one-hot labels for several applications including multi-class classification and semantic segmentation. For semantic segmentation, for example, a simple kernel smoothing of the segmentation mask is suggested to account for unreliable boundaries. Once this smooth label is obtained, the deep learning model is trained by minimizing the Kullback-Leibler (KL) divergence between the model output and the smooth noisy label. Label smoothing is a well-know trick for improving the test performance of deep learning models \cite{szegedy2016, muller2019}. The DLDL approach was improved by \cite{yi2019}, where the authors introduced a cross-entropy-based loss term to encourage closeness of estimated labels and the initial noisy labels and proposed a back-propagation method to iteratively update the initial label distributions as well.

\cite{ratner2016} used a generative model to model labeling of large datasets used in deep learning and proposed a label denoising method under this scenario. \cite{zhou2017b} proposed a GAN for removing label noise from synthetic data generated to train a CNN. This method was shown to be highly effective in removing label noise and improving the model performance. GANs were used to generate a training dataset with clean labels from an initial dataset with noisy labels by \cite{chiaroni2019}.

\subsection{Network architecture}

Several studies have proposed adding a ``noise layer" to the end of deep learning models. The noise layer proposed by \cite{sukhbaatar2014} is equivalent to multiplication with the transition matrix between noisy and true labels. The authors developed methods for learning this matrix in parallel with the network weights using error back-propagation. A similar noise layer was proposed by \cite{thekumparampil2018} for training a generative adversarial network (GAN) under label noise. \cite{sukhbaatar2014b} proposed methods for estimating the transition matrix from either a clean or a noisy dataset. Reductions of up to 3.5\% in classification error were reported on different datasets. A similar noise layer was proposed by \cite{goldberger2016}, where the authors proposed an EM-type method for optimizing the parameters of the noise layer. Importantly, the authors extended their model to the more general case where the label noise also depends on image features. This more complex case, however, could not be optimized with EM and a back-propagation method was exploited instead. \cite{bekker2016} used a combination of EM and error back-propagation for end-to-end training with a noise layer. \cite{jindal2016} suggested that aggressive dropout regularization (with a rate of $90 \%$) can improve the effectiveness of such noise layers.

Focusing on noisy labels obtained from multiple annotators, \cite{tanno2019} proposed a simple and effective method for estimating the correct labels and annotator confusion matrices in parallel with CNN training. The key observation was that, in order to avoid the ambiguity in simultaneous estimation of true labels and annotator confusion matrices, the traces of the confusion matrices had to be penalized. The entire model including the CNN weights and confusion matrices were learned via SGD. The method was shown to be highly effective in estimating annotator confusion matrices for various annotator types including inaccurate and adversarial ones. Improvements of 8-11\% in image classification accuracy were reported compared to the best competing methods.

A number of studies have integrated different forms of probabilistic graphical models into deep neural networks to handle label noise. \cite{xiao2015} proposed a graphical model with two discrete latent variables $y$ and $z$, where $y$ was the true label and $z$ was a one-hot vector of size 3 that denoted whether the label noise was zero, class-independent, or class-conditional. Two separate CNNs estimated $y$ and $z$, and the entire model was optimized in an EM framework. The method required a small dataset with clean labels. The authors showed significant gains compared with baseline CNNs in image classification from large datasets with noisy labels. \cite{vahdat2017} employed an undirected graphical model to learn the relationship between correct and noisy labels. The model allowed incorporation of domain-specific sources of information in the form of joint probability distribution of labels and hidden variables. Their method improved the classification accuracy of baseline CNNs by up to 3\% on three different datasets. For image classification, \cite{misra2016} proposed to jointly train two CNNs to disentangle the object presence and relevance in a framework similar to the graphical model-based methods described above. Model parameters and true labels were estimated using SGD. A more elaborate model was proposed by \cite{yao2018}, where an additional latent variable was introduced to model the trustworthiness of the noisy labels.

\subsection{Loss functions}

A large number of studies keep the model architecture, training data, and training procedures largely intact and only change the loss function \cite{izadinia2015}. \cite{ghosh2017} studied the conditions for robustness of a loss function to label noise for training deep learning models. They showed that mean absolute value of error, MAE, (defined as the $\ell_1$ norm of the difference between the true and predicted class probability vectors) is tolerant to label noise. This means that, in theory, the optimal classifier can be learned by training with basic error back-propagation. They showed that cross-entropy and mean square error did not possess this property. For a multi-class classification problem, denoting the vector of true and predicted probabilities with $p(y=j | x)$ and $\hat{p}(y=j | x)$, respectively, the cross-entropy loss function is defined as $L_{\text{CE}}= \sum_j p(y=j | x) \log \hat{p}(y=j | x)$. The MAE loss is defined as $L_{\text{MAE}}= \sum_j | p(y=j | x) - \hat{p}(y=j | x) |$. As opposed to cross-entropy that puts more emphasis on hard examples (desirable for training with clean labels), MAE tends to treat all data points more equally. However, a more recent study argued that because of the stochastic nature of the optimization algorithms used to train deep learning models, training with MAE down-weights difficult samples with correct labels, leading to significantly longer training times and reduced test accuracy \cite{zhang2018}. The authors proposed their own loss functions based on Box-Cox transformation to combine the advantages of MAE and cross-entropy. Similarly, \cite{wang2019c} analyzed the gradients of cross-entropy and MAE loss functions to show their weaknesses and advantages. They proposed an improved MAE loss function (iMAE) that overcame MAE's poor sample weighting strategy. Specifically, they showed that the $\ell_1$ norm of the gradient of $L_{\text{MAE}}$ with respect to the logit vector was equal to $4 p(y | x) (1- p(y | x))$, leading to down-weighting of difficult but informative data samples. To fix this shortcoming, they suggested to transform the MAE weights nonlinearly with a new weighting defined as $ \exp (T p(y | x)) (1- p(y | x))$, where the hyperparameter $T$ was set equal to 8 for training data with noisy labels. In image classification experiments on the CIFAR-10 dataset, compared with cross-entropy and MAE losses, their proposed iMAE loss improved the classification by approximately 1-5\% when label noise was low and up to 25\% when label noise was very high. In another experiment on person reidentification in video, iMAE improved the  mean average precision by 13\% compared with cross-entropy.

\cite{thulasidasan2019} proposed modifying the cross-entropy loss function to enable abstention. Their proposed modification allowed the model to abstain from making a prediction on some data points at the cost of incurring an abstention penalty. They showed that this policy could improve the classification performance on both random label noise as well as systematic data-dependent label noise. \cite{rusiecki2019} proposed a trimmed cross-entropy loss based on trimmed absolute value criterion. Their central assumption is that, with a well-trained model, data samples with wrong labels result in high loss values. Hence, their proposed loss function simply ignores the training samples with the largest loss values. Note that the central idea in \mbox{\cite{rusiecki2019}} (of down-weighting hard data samples) seems to run against many prevalent techniques in machine learning such as boosting \mbox{\cite{freund1999}}, hard example mining \mbox{\cite{shrivastava2016}}, and loss functions such as focal loss \mbox{\cite{lin2017}}, that steer the training process to focus on hard examples. This is because when the training labels are correct, data points with high loss values constitute the hard examples that the model has not learned yet. Hence, focusing on those examples generally helps improve the model performance. On the other hand, when there is significant label noise, assuming that the model has attained a decent level of accuracy, data points with unusually high loss values are likely to have wrong labels. This idea is not restricted to \mbox{\cite{rusiecki2019}} and it is an idea that is shared by many methods reviewed in this article. This paradigm shift is a good example of the dramatic effect of label noise on the machine learning methodology.

\cite{patrini2017} proposed two simple ways of improving the robustness of a loss function to label noise for training deep learning models. The proposed correction methods are based on the error confusion matrix $T$, defined as $T_{i,j}= p(\tilde{y}= e^j | y= e^i)$, where $\tilde{y}$ and $y$ are the noisy and true labels, respectively. Assuming $T$ is non-singular, one of the proposed correction strategies is $l_{\text{corr}}(\hat{p}(y|x))= T^{-1} l(\hat{p}(y|x))$. This correction is a linear weighting of the loss values for each possible label, where the weights, given by $T$, are the probability of the true label given the observed label. The authors name this correction method ``backward correction" because it is intuitively equivalent to going one step back in the noise process described by the Markov chain represented by $T$. The alternative approach, named forward correction, is based on correcting the model predictions and only applies to composite proper loss functions (\mbox{\cite{reid2010}}), which include cross-entropy. The corrected loss is defined as $l_{\text{corr}}(h(x))= l( T^{\text{T}} \psi^{-1} ((h(x)) )$, where $h$ is the vector of logits, and $\psi^{-1} $ is the inverse of the \textit{link function} for the loss function in consideration, which is the standard softmax for cross-entropy loss. The authors show that both these corrections lead to unbiased loss functions, in the sense that $ \forall x \, E_{\tilde{y}|x} \, l_{\text{corr}} = E_{y|x} \, l$. They also propose a method for estimating $T$ from noisy data and show that their methods lead to performance improvements on a range of computer vision problems and deep learning models. Similar methods have been proposed by \cite{hendrycks2018}, and \cite{boughorbel2018}, where it is suggested to use a small dataset with clean labels to estimate $T$. \cite{boughorbel2018} alternate between training on a clean dataset with a standard loss function and training on a larger noisy dataset with the corrected loss function. \cite{mnih2012} proposed a similar loss function based on penalizing the disagreement between the predicted label and the posterior of the true label.

\subsection{Data re-weighting}

Broadly speaking, these methods aim at down-weighting those training samples that are more likely to have incorrect labels. \cite{ren2018} proposed to weight the training data using a meta-learning approach. That method required a separate dataset with clean labels, which was used to determine the weights assigned to the training data with noisy labels. Simply put, it optimized the weights on the training samples by minimizing the loss on the clean validation data. The authors showed that this weighting scheme was equivalent to assigning larger weights to training data samples that were similar to the clean validation data in terms of both the learned features and optimization gradient directions. Experiments showed that this method improved upon baseline methods by 0.5\% and 3\% on CIFAR-10 and CIFAR-100 with only 1000 images with clean labels. More recently, \cite{wang2019b} proposed to re-weight samples by optimization gradient re-scaling. The underlying idea, again, is to give larger weights to samples that are easier to learn, hence more likely to have correct labels. Pumpout, proposed by \cite{han2018b}, is also based on gradient scaling. The authors propose two methods for identifying data samples that are likely to have incorrect lables. One of their methods is based on the assumption that data samples with incorrect labels are likely to display unusually high loss values. Their second method is based on the value of the backward-corrected loss \mbox{\cite{patrini2017}}; they suggest that the condition $\textbf{1}^T T^{-1} l(\hat{p}(y|x)) < 0$ indicates data samples with incorrect labels. For training data samples that are suspected of having incorrect labels, the gradients are scaled by $- \gamma$, where $0< \gamma <1$. In other words, they perform a scaled gradient \textit{ascent} on the samples with incorrect labels. In several experiments, including image classification with MNIST and CIFAR-10 datasets, they show that their method avoids fitting to incorrect labels and reduces the classification error by up to 40\%.

\cite{shen2019} proposed a training strategy that can be interpreted as a form of data re-weighting. In each training epoch, they remove a fraction of the data for which the loss is the largest, and update the model parameters to minimize the loss function on the remaining training data. This method assumes that the model gradually converges towards a good classifier such that the mis-labeled training samples exhibit unusually high loss values as training progresses. The authors proved that this simple approach learns the optimal model in the case of generalized linear models. For deep CNNs that are highly nonlinear, they empirically showed the effectiveness of their method on several image classification tasks. As in the case of this method, there is often a close connection between some of the data re-weighting methods and methods based on robust loss functions. \cite{shu2019} built upon this connection and developed it further by proposing to learn a data re-weighting scheme from data. Instead of assuming a pre-defined weighting scheme, they used a multi-layer perceptron (MLP) model with a single hidden layer to learn a suitable weighting strategy for the task and the dataset at hand. The MLP in this method is trained on a small dataset with clean labels. Experiments on datasets with unbalanced and noisy labels showed that the learned weighting scheme conformed with those proposed in other studies. Specifically, for data with noisy labels the model learned to down-weight samples with large loss functions, the opposite of the form learned for datasets with unbalanced classes. One can argue that this observation empirically justifies the general trend towards down-weighting training samples with large loss values when training with noisy labels.

A common scenario involves labels obtained from multiple sources or annotators with potentially different levels of accuracy. This is a heavily-researched topic in machine learning. A simple approach to tackling this scenario is to use expectation-maximization (EM)-based methods such as \cite{warfield2004,raykar2010} to estimate the true labels and then proceed to train the deep learning model using the estimated labels. \cite{khetan2017} proposed an iterative method, whereby model predictions were used to estimate annotator accuracy and then these accuracies were used to train the model with a loss function that properly weighted the label from each annotator. The model was updated via gradient descent, whereas annotator confusion matrices were optimized with an EM method. By contrast, \cite{tanno2019} estimated the network weights as well as annotator confusion matrices via gradient descent.

\subsection{Data and label consistency}

It is usually the case that the majority of the training data samples have correct labels. Moreover, there is considerable correlation among data points that belong to the same class (or the features computed from them). These correlations can be exploited to reduce the impact of incorrect labels. A typical example is the work of \cite{lee2019}, where the authors consider the correlation of the features learned by a deep learning model. They suggest that the features learned by various layers of a deep learning model on data samples of the same class should be highly correlated (i.e., clustered). Therefore, they propose training an ensemble of generative models (in the form of linear discriminant classifiers) on the features of the penultimate layer and possibly also other layers of a trained deep learning model. They show significant improvements in classification accuracy on several network architectures, noise levels, and datasets. On CIFAR-10 dataset, they report classification accuracy improvements of 3-20\%, with larger improvements for higher label noise levels, compared with a baseline CNN. On more difficult datasets such as CIFAR-100 and SVHN, smaller but still significant improvements of approximately 3-10\% are reported. Another example is the work of \cite{zhang2019}, where the authors proposed a method to leverage the multiplicity of data samples with the same (noisy) label in each training batch. All samples with the same label were fed into a light-weight neural network model that assigned a confidence weight to each sample based on the probability of it having the correct label. These weights were used to compute a representative feature vector for that class, which was then used to train the main classification model. Compared with other competing methods, 1-4\% higher classification accuracies were reprorted on several datasets. For face identification, \cite{speth2019} proposed feature embedding to detect data samples with incorrect labels. Their proposed verification framework used a multi-label Siamese CNN to embed a data point in a lower-dimensional space. The distance of the point to a set of representative points in this lower-dimensional space was used to determine whether the label was incorrect.

\cite{azadi2015} propose a method that they name ``auxiliary image regularization". Their method requires a small set of auxiliary images with clean labels in addition to the main training dataset with noisy labels. The core idea of auxiliary image regularization is to encourage representation consistency between training images (with noisy labels) and auxiliary images (with known correct labels). For this purpose, their proposed loss function includes a term based on group sparsity that encourages the features of a training image to be close to those of a small number of auxiliary images. Clearly, the auxiliary images should include good representatives of all expected classes. This method improved the classification accuracy by up to 8\% on ImageNet dataset. \cite{chen2019} proposed a manifold regularization technique that penalized the KL divergence between the class probability predictions of similar data samples. Because searching for similar samples in high-dimensional data spaces was challenging, they suggested using data augmentation to synthesize similar inputs. They reported 1-3\% higher classification accuracy compared with several alternative methods on CIFAR-10 and CIFAR-100. \cite{li2017b} proposed BundleNet, where multiple images with the same (noisy) labels were stacked together and fed as a single input to the network. Even though the authors do not provide a clear justification of their method and its difference with standard mini-batch training, they show empirically that their method improves the accuracy on image classification with noisy labels. \cite{wang2018b} used the similarity between images in terms of their deep features in an iterative framework to identify and down-weight training samples that were likely to have incorrect labels. Consistency between predicted labels and data (e.g., images or features) was exploited by \cite{reed2014}. The authors considered the true label as a hidden variable and proposed a model that simultaneously learned the relation between true and noisy labels (i.e., label noise distribution) and an auto-encoder model to reconstruct the data from the hidden variables. They showed improved performance in detection and classification tasks.


\subsection{Training procedures}

The methods in this category are very diverse. Some of them are based on well-known machine learning methods such as curriculum learning and knowledge distillation, while others focus on modifying the settings of the training pipeline such as learning rate and regularization.

Several methods based on curriculum learning have been proposed to combat label noise. Curriculum learning, first proposed by \cite{bengio2009}, is based on training a model with examples of increasing complexity or difficulty. In the method proposed by \cite{jiang2017}, an LSTM network called Mentor-Net provides a curriculum, in the form of weights on the training samples, to a second network called Student-Net. On CIFAR-100 and ImageNet with various label noise levels, their method improved the classification accuracy by up to 20\% and 2\%, respectively. \cite{guo2018} proposed another method based on curriculum learning, named CurriculumNet, for training a model from massive datasets with noisy labels. This method first clusters the training data in some feature space and identifies samples that are more likely to have incorrect labels as those that fall in low-density clusters. The data are then sequentially presented to the main CNN model to be trained. This technique achieved good results on several datasets including ImageNet. The Self-Error-Correcting CNN proposed by \cite{liu2017} is based on similar ideas; the training begins with noisy labels but as the training proceeds the network is allowed to change a sample's label based on a confidence policy that gives more weight to the network predictions with more training. 

\cite{li2017} adopted a knowledge distillation approach \cite{hinton2015} to train an auxiliary model on a small dataset with clean labels to guide the training of the main model on a large dataset with noisy labels. In brief, their approach amounts to using a pseudo-label, which is a convex combination of the noisy label and the label predicted by the auxiliary model. To reduce the risk of overfitting the auxiliary model on the small clean dataset, the authors introduced a knowledge graph based on the label transition matrix. \cite{reed2014} also proposed using a convex combination of the noisy labels and labels predicted by the model at its current training stage. They suggested that as the training proceeds, the model becomes more accurate and its predictions can be weighted more strongly, thereby gradually forgetting the original incorrect labels. \cite{zhong2019} used a similar approach for face identification. They first trained their model on a small dataset with less label noise and then fine-tuned it on data with stronger label noise using an iterative label update strategy similar to that explained above. Their method led to improvements of up to 2\% in face recognition accuracy. Following a similar training strategy, \cite{kohler2019} suggested that there is a point (e.g., a training epoch) when the model learns the true data features and is about to fit to the noisy labels. They proposed two methods, one based on the predictions on a clean dataset and another based on prediction uncertainty measures, to identify that stage in training. The output of the model at that stage can be used to fix the incorrect labels.

A number of studies have proposed methods involving joint training of more than one model. For example, one work suggested simultaneously training two separate but identical networks with random initialization, and only updating the network parameters when the predictions of the two networks differed \cite{malach2017}. The idea is that when training with noisy labels, the model starts by learning the patterns in data samples with correct labels. Later in training, the model will struggle to overfit to samples with incorrect labels. The proposed method hopes to reduce the impact of label noise because the decision as to whether or not to update the model is made based on the predictions of the two models and independent of the noisy label. In other words, on data with incorrect labels both models are likely to produce the same prediction, i.e., they will predict the correct label. On easy examples with correct labels, too, both models will make the same (correct) prediction. On hard examples with correct labels, on the other hand, the two models are more likely to disagree. Hence, with the proposed training strategy, the data samples that will be used in later stages of training will shrink to the hard data samples with correct labels. This strategy also improves the computational efficiency since it performs many updates at the start of training but avoids unnecessary updates on easy data samples once the models have sufficiently converged to predict the correct label on those samples. This idea was developed into co-teaching \cite{han2018}, whereby the two networks identified label-noise-free samples in their mini-batches and shared the update information with the other network. The authors compare their method with several state of the art techniques including Mentor-Net (\mbox{\cite{jiang2017}}. Their method outperformed competing methods in most experiments, while narrowly underperforming in some experiments. Co-teaching was further improved in \cite{yu2019}, where the authors suggested to focus the training on data samples with lower loss values in order to reduce the risk of training on data with incorrect labels. Along the same lines, \cite{li2019} proposed a meta-learning objective that encouraged consistent predictions between a student model trained on noisy labels and a teacher model trained on clean labels. The goal was to train the student model to be tolerant to label noise. Towards this goal, artificial label noise was added on data with correct labels to train the student model. The student model was encouraged to be consistent with the teacher model using a meta-objective in the form of the KL divergence between prediction probabilities. Their method outperformed several competing methods by 1-2\% on CIFAR-10 and Clothing1M datasets.

Experiments in \cite{chen2019b} showed that co-teaching was less effective as the label noise increased. Instead, the authors showed that selecting the data samples with correct labels using cross-validation was more effective. In their proposed approach, the training data was divided into two folds. The model was iteratively trained on one fold and tested on the other. Data samples for which the predicted and noisy labels agreed were assumed to have the correct label and were used in the next training epoch. One study proposed to learn the network parameters by optimizing the joint likelihood of the network parameters and true labels \cite{tanaka2018}. Compared with standard training with cross-entropy loss, this method improved the classification accuracy on CIFAR-10 by 2\% with low label noise rate to 17\% when label noise rate was very high.

Some studies have suggested modifying the learning rate, batch size, or other settings in the training methodology. For example, for applications where multiple datasets with varying levels of label noise are available, \cite{song2015} have proposed training strategies in terms of the order of using different datasets during training and proper learning rate adjustments based on the level of label noise in each dataset. Assuming that separate clean and noisy datasets are available, the same study has shown that using different learning rates for training with noisy and clean samples can improve the performance. It has also shown that the optimal ordering of using the two datasets (i.e., whether to train on the noisy dataset or the clean dataset first) depends on the choice of the learning rate. It has also been suggested that when label noise is strong, the effective batch size decreases, and that batch size should be increased with a proper scaling of the learning rate \cite{rolnick2017}. \cite{sukhbaatar2014b} proposed to include samples from a noisy dataset and a clean dataset in each training mini-batch, giving higher weights to the samples with clean labels.

\textit{Mixup} is a less intuitive but simple and effective method \cite{zhang2017}. It synthesizes new training data points and labels via a convex combination of pairs of training data points and their labels. More specifically, given two randomly selected training data and label pairs $(x_i, y_i)$ and $(x_j, y_j)$, a new training data point and label are synthesized as $\tilde{x}= \lambda x_i+ (1-\lambda) x_j$ and $\tilde{y}= \lambda y_i+ (1-\lambda) y_j$, where $\lambda \in [0,1]$ is sampled from a beta distribution. Although mixup is known primarily as a data augmentation and regularization strategy, it has been shown to be remarkably effective for combatting label noise. Compared with basic emprirical risk minimization on CIFAR-10 dataset with different levels of label noise, mixup reduced the classification error by 6.5-12.5\%. The authors argue that the reason for this behavior is because interpolation between datapoints makes memorization on noisy labels, as observed in \mbox{\cite{zhang2016}}, more difficult. In other words, it is easier for the network to learn the linear iterpolation between datapoints with correct labels than to memorize the interploation when labels are incorrect. The same idea was successfully used in video classification by \cite{ostyakov2018}.

For object boundary segmentation, two studies proposed to improve noisy labels in parallel with model training \cite{yu2018,acuna2019}. This is a task for which large datasets are known to suffer from significant label noise and model performance to be very sensitive to label noise. Both methods consider the true boundary as a latent variable that is estimated in an alternating optimization framework in parallel with model training. One major assumption in \cite{acuna2019} is the preservation of the length of the boundary during optimization, resulting in a bipartite graph assignment problem. In \cite{yu2018}, a level-set formulation was introduced instead, providing much higher flexibility in terms of the shape and length of the boundary while preserving its topology. Both studies compared their methods with baseline CNNs in terms of F-measure for object edge detection and report impressive improvements. In particular, \mbox{\cite{acuna2019}} improved upon their baseline CNN by 2-5\% on segmentation of different objects. Similarly, \mbox{\cite{yu2018}} reported improvements of 1-17\% compared to a baseline CNN.


\section{Deep learning with noisy labels in medical image analysis}
\label{label_noise_in_MIA}

In this section, we review studies that have addressed label noise in training deep learning models for medical image analysis. We use the same categorization as in the previous section.

\subsection{Label cleaning and pre-processing}

For classification of thoracic diseases from chest x-ray scans, \mbox{\cite{pham2019}} used label smoothing to handle noisy labels. They compared their label smoothing method with simple methods such as ignoring data samples with noisy labels. They found that label smoothing can lead to improvements of up to 0.08 in the area under the receiver operating characteristic curve (AUC).

\subsection{Network architectures}

The noise layer proposed by \cite{bekker2016}, reviewed above, was used for breast lesion detection in mammograms by \cite{dgani2018} and slightly improved the detection accuracy.

\subsection{Loss functions}

To train a network to segment virus particles in transmission electron microscopy images using original annotations that consisted of only the approximate center of each virus, \cite{matuszewski2018} dilated the annotations with a small and a large structuring element to generate noisy masks for foreground and background, respectively. Consequently, parts of the image in the shape of the union of rings were marked as uncertain regions that were ignored during training. The Dice similarity and intersection-over-union loss functions were modified to ignore those regions. Promising results were reported for both loss functions. \mbox{\cite{rister2018}} showed that for segmentation of abdominal organs in CT images from noisy training annotations, the intersection-over-union (IOU) loss consistently outperformed the cross-entropy loss. The mean DSC achieved with the IOU loss was 1-13\%  higher than the DSC achieved with the cross-entropy loss.

\subsection{Data re-weighting}

\cite{le2019} used a data re-weighting method similar to that proposed by \cite{ren2018} to deal with noisy annotations in pancreatic cancer detection from whole-slide digital pathology images. They trained their model on a large corpus of patches with noisy labels using weights computed from a small set of patches with clean labels. This strategy improved the classification accuracy by $10 \%$ compared with training on all patches with clean and noisy labels without re-weighting. For skin lesion classification in dermoscopy images with noisy labels, \cite{xue2019} used a data re-weighting method that amounted to removing data samples with high loss values in each training batch. This method, which is similar to some of the methods reviewed above such as the method of \cite{shen2019}, increased the classification accuracy by $2-10 \%$, depending on the label noise level.

For segmentation of heart, clavicles, and lung in chest radiographs,  \cite{zhu2019} trained a deep learning model to detect incorrect labels. This model assigned a weight to each sample in a training batch, aiming to down-weight samples with incorrect labels. The main segmentation model was trained in parallel using a loss function that made use of these weights. A pixel-wise weighting was proposed by \cite{mirikharaji2019} for skin lesion segmentation from highly inaccurate annotations. The method needed a small dataset with correct segmentations alongside the main, larger, dataset with noisy segmentations. For each training image with noisy segmentation, a weight map of the same size was considered to indicate the pixel-wise confidence in the accuracy of the noisy label. These maps were updated in parallel with network parameters with alternating optimization. The authors proposed to optimize the weights on the images in the noisy dataset by reducing the loss on the clean dataset. In essence, the weight on a pixel is increased if that leads to a reduction in the loss on the clean dataset. If increasing the weight on a pixel increases the loss on the clean dataset, that weight is set to zero because the label for that pixel is probably incorrect.

\subsection{Data and label consistency}

For segmentation of the left atrium in MRI from labeled and unlabeled data, \mbox{\cite{yu2019b}} proposed training two separate models: a teacher model that produced noisy labels and label uncertainty maps on unlabeled images, and a student model that was trained using the generated noisy labels while taking into account the label uncertainty. The student model was trained to make correct predictions on the clean dataset and to be consistent with the teacher model on noisy labels with uncertainty below a threshold. The teacher model was updated in a moving average scheme involving the weights of the student model.

\subsection{Training procedures}


For bladder, prostate, and rectum segmentation in MRI, \cite{nie2018} trained a model on a dataset with clean labels and used it to predict segmentation masks for a separate unlabeled dataset. In parallel, a second model was trained to estimate a confidence map to indicate the regions where the predicted labels were more likely to be correct. The confidence maps were used to sample the unlabeled dataset for additional training data for the main model. Improvements of approximately 3\% in Dice similarity coefficient (DSC) were reported.

\cite{min2018} employed the ideas proposed by \cite{malach2017} to develop label-noise-robust methods for medical image segmentation. As we reviewed above, the main idea in the method of \cite{malach2017} was to jointly train two separate models and update the models only on the data samples on which the predictions of the two models differed. Instead of considering only the final layer predictions, \cite{min2018} introduced attention modules at various depths in the networks to use the gradient information at different feature maps to identify and down-weight samples with incorrect labels. They reported promising results for cardiac and glioma segmentation in MRI.

For cystic lesion segmentation in lung CT, \cite{zhang2018b} generated initial noisy segmentations using unsupervised K-means clustering. These segmentations were used to train a CNN. Assuming that the CNN was more accurate than K-means, CNN predictions were used as the training labels for the next epoch. This process was repeated, generating new labels at the end of each training epoch. Experiments showed that the final trained CNN achieved significantly higher segmentation accuracy compared with the K-means method used to generate the initial segmentations. A rather similar method was used for classification of aortic valve malfunctions in MRI by \cite{fries2019}. Using a small dataset of expert-annotated images, simple classifiers based on intensity and shape features were developed. Subsequently, a factor graph-based model was trained to estimate the classification accuracies of these classifiers and to generate pseudo-ground-truth labels on a massive unlabeled dataset. This dataset was then used to train a deep learning classifier. This model significantly outperformed models trained on a small set of expert-labeled images.

\section{Experiments}
\label{experiments}

In this section, we present our experiments on three medical image datasets with noisy labels, in which we explored several methods that we implemented, adapted, or developed to analyze and reduce the effect of label noise. Our experiments represent three different machine learning problems, namely, detection, classification, and segmentation. The three datasets associated with these experiments represent three different noise types, namely, label noise due to systematic error by a human annotator, label noise due to inter-observer variability, and error/noise in labels generated by an algorithm (Figure~\ref{fig:methods_schematics}). In developing and comparing techniques, our goal was not to achieve the best, state-of-the-art results in each experiment, as that would have required careful design of network architectures, data pre-processing, and training procedures for each problem. Instead, our goal was to show the effects of label noise and the relative effectiveness, merits, and shortcomings of potential methods on common label noise types in medical image datasets. 

\subsection{Brain lesion detection and segmentation}
\label{tsc_experiment}

\subsubsection{Data and labels}

\begin{figure}[htb]
\begin{minipage}[b]{1.0\linewidth}
  \centering
  \centerline{\includegraphics[width=8.0cm]{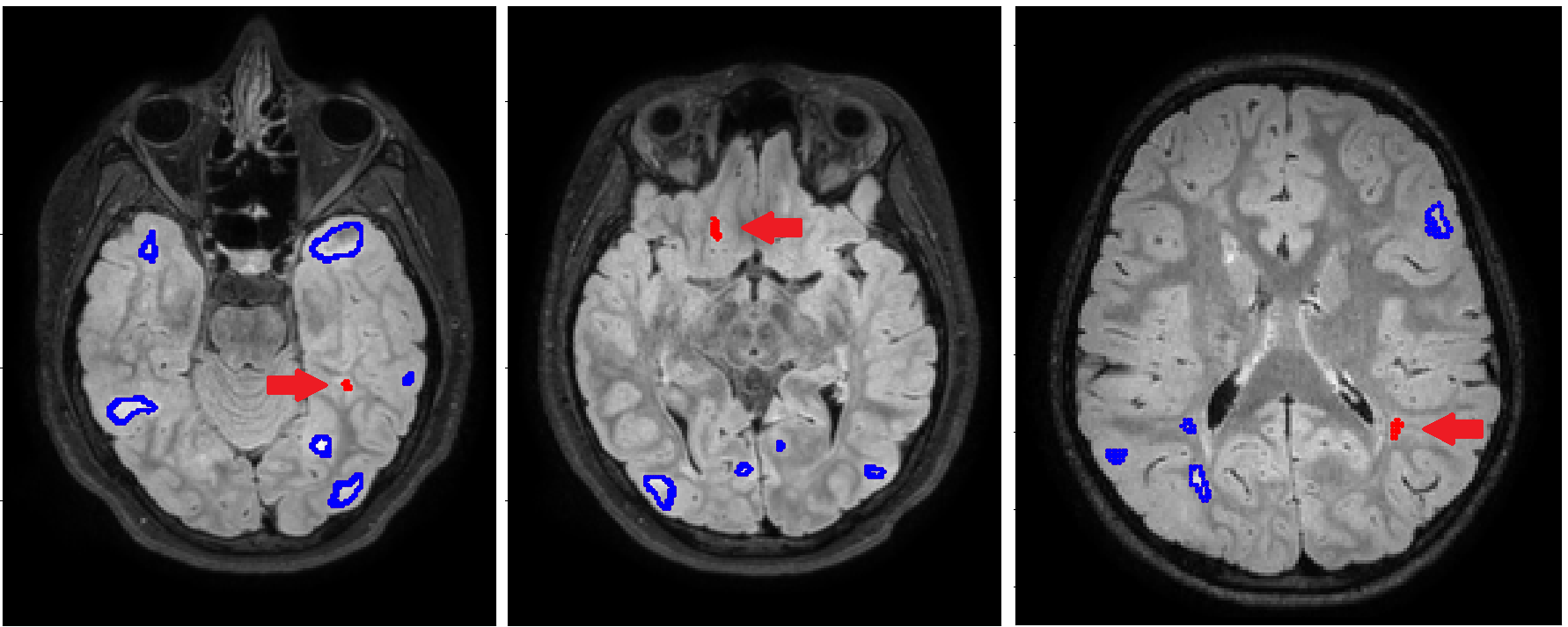}}
\end{minipage}
\caption{The FLAIR images from three TSC subjects and the lesions that were detected (in blue) and missed (in red) by an experienced annotator in the first reading.}
\label{fig:example_scan}
\end{figure}

We used 165 MRI scans from 88 tuberous sclerosis complex (TSC) subjects. Each scan included T1, T2, and FLAIR images. An experienced annotator segmented the lesions in these scans. We then randomly selected 12 scans for accurate annotation and assessment of label noise. Two annotators jointly reviewed these scans in four separate sessions to find and fix missing or inaccurate annotations. The last reading did not find any missing lesions in any of the 12 scans. Example scans and their annotations are shown in Figure \ref{fig:example_scan}. We used these 12 scans and their annotations for evaluation only. We refer to these scans as ``the clean dataset". We used the remaining 153 scans and their imperfect annotations for training. These are referred to as ``the noisy dataset".  

In the 12 scans in the clean dataset, 306 lesions were detected in the first reading and 68 lesions in the followup readings, suggesting that approximately 18\% of the lesions were missed in the first reading. Annotation error can be modeled as a random variable, in the sense that if the same annotators annotate the same scan a second time (with some time interval) they may not make the same errors. Nonetheless, our analysis shows that smaller or fainter lesions were more likely to be missed. Specifically, Welchs t-tests showed that the lesions that had been missed in the first reading were less dark on the T1 image ($p\!<\!0.001$), smaller in size ($p\!<\!0.001$), and farther away from the closest lesion ($p\!=\!0.004$), compared with lesions that were detected in the first reading. Therefore, in this application the intrinsic limitation of human annotator attention results in systematic errors (noise) in labels.

\subsubsection{Methods}
\label{tsc_experiment_methods}

For the joint detection and segmentation of lesions in this application, we used a baseline CNN similar to the 3D U-Net \cite{cciccek2016}. This CNN included four convolutional blocks in each of the contracting and expanding parts. The first convolutional block included 14 feature maps, which increased by a factor of 2 in subsequent convolutional blocks, resulting in the coarsest convolutional block with 112 feature maps. Each convolutional block processed its feature maps with additional convolutional layers with residual connections. All convolutional operations were followed by ReLU activation. The CNN worked on blocks of size $64^3$ voxels and it was applied in a sliding-window fashion to process an image. In addition, since this application can be regarded as a detection task, we also used a method based on Faster-RCNN \mbox{\cite{ren2015}}, where we used a 3D U-Net architecture for the backbone of this method. To train Faster-RCNN, we followed the training methodology of \mbox{\cite{ren2015}}, but made changes to adapt it to 3D images. Based on the distribution of lesion size in our data, we used five different anchor sizes and three different aspect ratios, for a total of 15 anchors in Faster-RCNN. The smallest and largest anchors were $3 \times 3 \times 7 \text{ mm}^3$ and $45 \times 45 \times 61 \text{ mm}^3$, respectively. Our evaluation was based on two-fold subject-wise cross-validation, each time training the model on data from approximately half of the subjects and testing on the remaining subjects. 
Following the latest recommendations in the literature on lesion detection applications~\cite{carass2017,commowick2018,hashemi2019}, our main evaluation criterion was lesion-count F1 score; but since this is considered a joint segmentation and detection task, we also computed DSC when applicable (i.e., for the 3D U-Net). It is noteworthy that due to the criteria that are used in diagnosis/prognosis and disease modifying treatments, lesion-count measures such as lesion-count F1-score have been considered more appropriate performance measures for lesion detection and segmentation algorithms compared to DSC \cite{commowick2018,hashemi2019}.

The methods developed, implemented, and compared in this task include:

\begin{itemize}

\item{Faster-RCNN trained on noisy labels.}

\item{Faster-RCNN trained on clean data. Same as the above, but evaluated using two-fold cross-validation on the clean data}.

\item{Faster-RCNN trained with MAE loss \mbox{\cite{ghosh2017}.}}

\item{3D U-Net CNN} trained on noisy labels with DSC loss.

\item{3D U-Net CNN trained on clean data.} Same as the above, but evaluated using two-fold cross-validation on the clean data.

\item{3D U-Net CNN trained with MAE loss \cite{ghosh2017}.}

\item{3D U-Net CNN trained with iMAE loss \cite{wang2019c}.}

\item{3D U-Net CNN with data re-weighting.} In this method, we ignored data samples with very high loss values. We kept the mean and standard deviation of the losses of the 100 most recent training samples. If the loss for a training sample was higher than 1.5 standard deviations of the mean, the network weights were not updated on that sample. To the best of our knowledge, such a method has not been proposed for brain lesion detection/segmentation prior to this work.

\item{Iterative label cleaning.} This is a novel technique that we have developed for this application. We first trained a random forest classifier to distinguish the true lesions missed by the annotator from the false positive lesions in CNN predictions. This classification was based on six lesion features: mean image intensity in T1, T2, and FLAIR, lesion size, distance to the closest lesion, and mean prediction uncertainty, where uncertainty was computed using the methods of \cite{kendall2017}. Then, during training of the CNN on the noisy dataset, after each training epoch the random forest classifier was applied on the CNN-detected lesions that were not present in the original noisy labels. Lesions that were classified as true lesions were added to the noisy labels. Hence, this method iteratively improved the noisy labels in parallel with CNN training. 

\end{itemize}

\subsubsection{Results}

As shown in Table \mbox{\ref{table:tsc_results}}, 3D U-Net achieved higher detection accuracy than Faster-RCNN. Since our focus is on label noise, we discuss the results of experiments with each of these two networks independently. For 3D U-Net, both MAE and iMAE loss functions resulted in lower lesion-count F1 score and DSC, compared with the baseline CNN trained with a DSC loss. However, both MAE and iMAE have been proposed as improvements to the cross-entropy. With a cross-entropy loss, our CNN achieved performance similar to iMAE. Interestingly, for Faster-RCNN, compared with the baseline that was trained with the cross-entropy loss, using the MAE loss did improve the lesion-count F1 score by 0.041. This indicates that such loss functions, initially proposed for classification and detection tasks, may be more useful for lesion detection than for lesion segmentation applications. The data re-weighting method resulted in lesion-count F1 score and DSC that were substantially higher than the baseline CNN. Moreover, iterative label cleaning achieved much higher lesion-count F1 score and DSC than the baseline and outperformed the data re-weighting method too. The increase in the lesion-count F1 score shows that iterative label cleaning improves detection of small lesions. The increase in DSC is also interesting and less expected since small lesions account for a small fraction of the entire lesion volume, which greatly affects the DSC. We attribute the increase in DSC to a better training of the CNN with improved labels. In other words, improving the labels by detecting and adding small lesions helped learning a better CNN that performed better on segmenting larger lesions as well. Comparing the first and the second rows of Table \ref{table:tsc_results} shows that training on the clean dataset achieved results similar to training on the noisy dataset that included an order of magnitude larger number of scans. A similar observation was made for Faster-RCNN, where the lesion-count F1 score increased by 0.012 when trained on the clean dataset. This shows that in this application a small dataset with clean labels can be as good as a large dataset with noisy labels. In creating our clean dataset, we had to limit ourselves to a small number (12) of scans due to limited annotator time. It is likely that the results could further improve with a larger clean dataset.

\begin{table*}[!htb]
\footnotesize
  \begin{center}
    \begin{tabular}{ L{7.5cm}  C{1.2cm} C{3.2cm}}
\hline
Method & DSC & lesion-count F1 score  \\ \hline
Faster-RCNN trained on noisy labels & - &  0.541 \\
Faster-RCNN trained on clean data & - & 0.553 \\
Faster-RCNN trained with MAE loss \mbox{\cite{ghosh2017}} & - & 0.582 \\
3D U-Net CNN    & 0.584 & 0.747  \\
3D U-Net CNN trained on clean data    & 0.578 & 0.743  \\
3D U-Net CNN trained with MAE loss  & 0.541 & 0.695  \\ 
3D U-Net CNN trained with iMAE loss   & 0.485 & 0.657  \\ 
3D U-Net CNN with data re-weighting   & 0.600 & 0.802  \\ 
3D U-Net with Iterative label cleaning & \textbf{0.605} & \textbf{0.819}  \\ 
\hline
\end{tabular}
  \end{center}
 \caption{\small{Performance metrics (DSC and lesion-count F1 score) obtained in the experiment on TSC brain lesion detection using different techniques listed in Section~\ref{tsc_experiment_methods} compared with the baseline models trained with noisy labels (i.e., Faster-RCNN trained on noisy labels, and 3D U-Net CNN) and baseline models trained on clean data (i.e., Faster-RCNN trained on clean data, and 3D U-Net trained on clean data). The best performance metric value (in each column) has been highlighted in bold. The results show that in this application methods based on data re-weighting and iterative label cleaning substantially improved the performance of the CNNs trained with noisy labels. The best results in terms of both the DSC and the lesion-count F1 score were obtained from our 3D U-Net with iterative label cleaning.}}
  \label{table:tsc_results}
\end{table*}

\subsection{Prostate cancer digital pathology classification}
\label{pathology_experiment}

\subsubsection{Data and labels}

\begin{figure}[htb]
\begin{minipage}[b]{1.0\linewidth}
  \centering
  \centerline{\includegraphics[width=8.0cm]{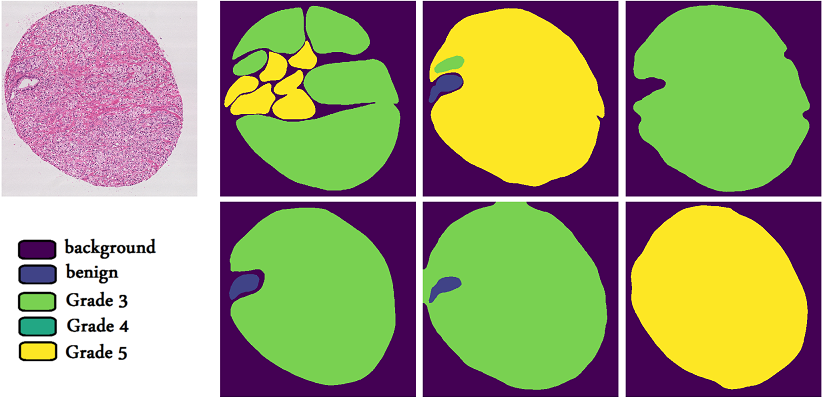}}
  \end{minipage}
\caption{An example TMA core image (top left) and annotations from six pathologists (right) with four labels: benign and Gleason cancer grades 3, 4, and 5.}
\label{fig:tma_core}
\end{figure}

We use the data from Gleason2019 challenge. The goal of the challenge is to classify prostate tissue micro-array (TMA) cores as one of the four classes: benign and cancerous with Gleason grades 3, 4, and 5. Data collection and labeling have been described by \cite{nir2018}. In summary, TMA cores have been classified in detail (i.e., pixel-wise) by six pathologists independently. The Cohen's kappa coefficient for the general pathologists on this task is approximately between 0.40 and 0.60 \cite{allsbrook2001,nir2018}, where a value of 0.0 indicates chance agreement and 1.0 indicates perfect agreement. The inter-observer variability also depends on experience \cite{allsbrook2001}; pathologists who labeled this dataset had different experience levels, ranging from 1 to 27 years. Hence, this is a classification problem and label noise is caused by inter-observer variability due to the subjective nature of grading. An example TMA core and pathologists' annotations are shown in Figure \ref{fig:tma_core}.

\subsubsection{Methods}

We used a MobileNet CNN architecture, which had been shown to be a good choice for this application by \cite{arvaniti2018,karimi2019deep} and used patches of size $768 \times 768$ pixels at 40X magnification as suggested by \cite{arvaniti2018}. The main feature of MobileNets is the use of separable convolutional filters, which replace a 3D convolution with a 2D depth-wise convolution (applied separately to each of the input feature maps) followed by a 1D convolution to combine these depth-wise convolutions. Our network had a depth of 7 convolutional blocks. The first block had 16 feature maps. The number of feature maps increased by a factor of 2 in each subsequent block, while reducing their size by a factor of 2 in each dimension. The output of the final convolutional block was flattened and passed through two fully connected layers. All convolutional and fully connected layers were followed by ReLU activations. 

An important consideration in this application was how to divide the labels from different pathologists for training and test stages. For most of our experiments, we used the labels from all six pathologists to estimate the ground truth labels on the test data using the Simultaneous Truth and Performance Level Estimation (STAPLE) \mbox{\cite{warfield2004}}. Our justification here is that, given the high inter-observer variability, this would be our best estimate of the ground truth. For these experiments, we followed a 5-fold cross-validation. Each time, we trained the CNN on 80\% of the TMA cores and their labels from the six pathologists and then evaluated the trained CNN on the STAPLE-estimated labels of the remaining 20\% of the cores. However, from the viewpoint of separation between test and train data, this may not be the best approach. Therefore, we performed another set of experiments, where we used the labels from three of the pathologists for training and used STAPLE-estimated ground truth from the other three pathologists on the test set for evaluation. For this set of experiments, too, we followed a 5-fold cross-validation. However, we repeated these experiments twice, each time using labels from three of the pathologists for training. Therefore, each TMA core was tested on twice. We report the average of the two results. Below, we denote the results for this set of experiments with ``3-3".

We compared the CNN predictions with the estimated truth by computing the classification accuracy and AUC for 1) distinguishing cancerous (Gleason grades 3-5) from benign tissue, and 2) in separating high-grade (Gleason grades 4 and 5) from low-grade (Gleason grade 3) cancer. In addition, we report the percentage of large classification errors, which we define as when the predicted class is 2 or 3 classes away from the true class, such as when Gleason grade 5 is classified as benign or Gleason grade 3. The compared methods were the following:

\begin{itemize}

\item{Single pathologist.} We used the label provided by one of the pathologists only, ignoring the labels provided by the others. We repeated this for all six pathologists.

\item{Majority vote.} We computed the pixel-wise majority vote and used that for training.

\item{STAPLE.} We used STAPLE to compute a pixel-wise label and used that for training.

\item{STAPLE + iMAE loss.} Similar to the above, but instead of the cross-entropy loss, we used the iMAE loss \cite{wang2019c}.

\item{Minimum-loss label.} On each training patch, we computed the loss on labels provided by each of the six pathologists and selected the one with the smallest loss for error back-propagation. To the best of our knowledge, this method has not been proposed previously for this application. 

\item{Annotator confusion estimation.} We used the method of \cite{tanno2019}, which we reviewed above. This method estimates the labeling patterns of the annotators in parallel with the training of the CNN classification model.

\end{itemize}

\subsubsection{Results}

\begin{figure*}[htb]
\begin{minipage}[b]{1.0\linewidth}
  \centering
  \centerline{\includegraphics[width=16.0cm]{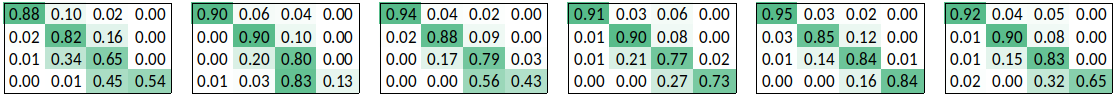}}
  \end{minipage}
\caption{Examples of the annotator confusion matrices estimated by the method of \cite{tanno2019} on the prostate cancer digital pathology data. In each matrix, rows represent for the estimated true label and columns represent the annotator's labels. Classes are in this order: benign and Gleson grades 3-5.}
\label{fig:confusion}
\end{figure*}

Table \ref{table:pathology} summarizes our results. The first row shows the average of results when using the labels from one of the six pathologists. Comparing this row with the second and third rows and the row denoted as ``STAPLE (3-3)" shows significant improvements due to using labels from multiple experts. Using the iMAE loss considerably improved the accuracy, especially for classifying cancerous from benign tissue. The \textit{minimum-loss label} method also improved the classification accuracy. The iMAE loss and minimum-loss label method are based on a similar philosophy: to combat label noise, data samples with unusually high loss values should be down-weighted because they are likely to have incorrect labels. While the iMAE loss down-weights the effect of such data samples, minimum-loss label aims at ignoring incorrect labels by using only the label with the lowest loss for each data sample. The iMAE loss performed better on classifying cancerous vs. benign tissue, whereas the minimum-loss label method performed better than the iMAE loss on classifying high-grade vs. low-grade cancer. This may be because the minimum-loss label method has a more aggressive label denoising policy and label noise (manifested as inter-pathologist disagreement) is known to be higher for high-grade vs. low-grade annotation compared with benign vs. cancerous annotation \cite{gulshan2016,nir2018}. Annotator confusion estimation also significantly improved the accuracy compared with the baseline methods. It can be argued that it is the best among the compared methods, as it achieved the best accuracy on high-grade vs. low-grade classification and close to the best accuracy on cancerous vs. benign classification. It also displayed the lowest rate of large classification errors at 1\%. The estimated annotator confusion matrices are shown in Figure \ref{fig:confusion}, which show that the pathologists had a low disagreement for benign vs. cancerous classification but relatively higher disagreement in cancer grading.

Overall, the results when labels from separate pathologists were used for training and test stages, presented in the last three rows of the table, showed similar conclusions. Specifically, using iMAE loss or modeling annotator accuracies led to better results than with cross-entropy loss and much better than when labels from a single expert were used. However, the results were worse than when labels from all six pathologists were used for training and for estimating the truth for the test set, especially for classifying high-grade versus low-grade cancer. We attribute this partly to the high inter-observer variability, which makes the estimated truth more accurate when labels from all six pathologists are used. However, this can also be because using labels from all six pathologists for training and test stages causes some overfitting that is avoided when labels from separate pathologists are used for training and test. 

\begin{table*}[!htb]
\footnotesize
  \begin{center}
    \begin{tabular}{ L{4.8cm} C{1.2cm} C{1.2cm} C{1.2cm} C{1.2cm} C{3.0cm} }
\hline
 &  \multicolumn{2}{c}{Cancerous vs. benign} &  \multicolumn{2}{c}{High-grade vs. low-grade} & \multirow{2}{*}{\parbox{2.5cm}{Percentage of large classification errors}}\\ 
Method & accuracy & AUC & accuracy &  AUC &  \\
\hline
Single pathologist & 0.80 & 0.78 & 0.65 & 0.61 & 0.07   \\
Majority vote & 0.86 & 0.87 & 0.73 & 0.74 & 0.03   \\
STAPLE & 0.84 & 0.86 & 0.73 & 0.72 & 0.03   \\
STAPLE + iMAE loss & \textbf{0.93} & 0.91 & 0.76 & 0.79 & 0.03   \\
Minimum-loss label & 0.88 & 0.88 & \textbf{0.80} & \textbf{0.82} & 0.03   \\
Annotator confusion estimation & 0.92 & \textbf{0.93} & \textbf{0.80} & \textbf{0.82} & \textbf{0.01}   \\ 
STAPLE (3-3) & 0.86 & 0.86 & 0.69 & 0.70 & 0.02   \\
STAPLE + iMAE loss (3-3) & 0.90 & 0.88 & 0.75 & 0.78 & 0.02   \\
Annotator confusion estimation (3-3) & 0.90 & 0.88 & 0.73 & 0.76 & 0.03 \\
\hline
\end{tabular}
  \end{center}
 \caption{\small{Results of the experiment on prostate cancer digital pathology classification using different methods. The highest accuracy in each classification task (column) has been highlighted in bold text.}}
  \label{table:pathology}
\end{table*}

\subsection{Fetal brain segmentation in diffusion-weighted MRI}
\label{segmentation_experiment}

\subsubsection{Data and labels}

\begin{figure}[htb]
\begin{minipage}[b]{1.0\linewidth}
  \centering
  \centerline{\includegraphics[width=8.0cm]{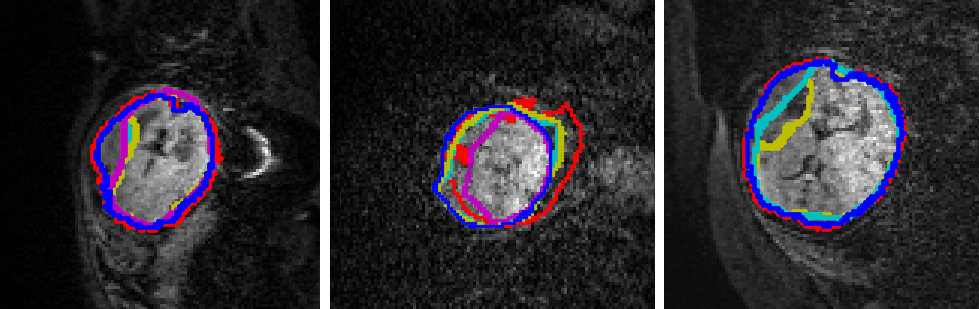}}
\end{minipage}
\caption{Examples of DW-MR fetal brain images along with manual segmentation (blue) and several noisy segmentations (other colors).}
\label{fig:dwi_masks}
\end{figure}

A total of 2562 diffusion weighted (DW) MR images from 65 fetuses (between 12 and 96 images from each fetus) were used in this experiment. One image from each fetus was manually segmented by two experienced annotators. We refer to these as ``clean data" and use them for evaluation. For the remaining 2497 images (between 11 and 95 images from each fetus), we generated approximate (i.e., noisy) segmentations using different methods. \textbf{Method 1:} these fetuses had reconstructed T2-weighted MR images with accurate brain segmentations, which we could transfer to the DW images via image registration. \textbf{Method 2:} we developed an algorithm based on intensity thresholding and morphological operations to synthesize approximate segmentations. This algorithm sometimes generated very inaccurate segmentations, which were detected by computing the DSC between them and the segmentation masks from the T2 image. If this DSC was below a threshold, we replaced the synthesized segmentation with that from the T2 image. This threshold and the parameters of the algorithm can be tuned to generate noisy segmentations with different accuracy levels. \textbf{Method 3:} we used a level set method to generate noisy labels. The level set method needs a seed to initialize the segmentation. In one variation of this method, we generated the seed by eroding the segmentation obtained from the T2 image mentioned above (Method 1). This resembles a semi-automatic method, where the level set method is initialized manually. In another, fully-automatic, variation of this method we used the rough segmentations generated by Method 2, after erosion, to initialize the level set method. After every 50 training epochs, the current CNN predictions were used to initialize the level set method and new training labels were generated. To assess the accuracy of the synthesized segmentations for each method and parameter settings, we applied that method on the 65 images in the clean dataset and computed the DSC between the synthesized and manual segmentations. Figure \ref{fig:dwi_masks} shows example scans from the clean dataset and several noisy segmentations.

\subsubsection{Methods}

We trained a CNN, similar to 3D U-Net for experiments in this section. This architecture included four convolutional blocks in each of its contracting and expanding parts. The first block extracted 10 feature maps from the image. The number of feature maps increased by a factor of 2 in subsequent convolutional blocks. Each convolutional block included two standard convolutional layers with a residual connection. Similar to the other networks used in this work, a ReLU activation was used after each convolutional operation. We adopted a five-fold cross-validation strategy for all experiments in this section. The cross-validation was subject-wise, meaning that no scans from the test subjects were used for training. The compared training methods were:

\begin{itemize}

\item{Baseline CNN.} 

\item{Baseline CNN trained with MAE loss.}

\item{Dual CNNs with iterative label update.} This is a novel method that we propose for fetal brain segmentation for the first time. We trained two CNNs, with the same architecture as the baseline CNN, but with 0.80 and 1.25 times the number of feature maps as the baseline CNN to encourage diversity. The CNNs were first trained on the initial noisy labels. Subsequently, they were used to predict segmentations on the images with noisy labels. In an iterative framework, first each CNN was trained using the labels predicted by the other CNN or the noisy label, whichever resulted in a lower loss. Then, at the end of each training epoch, each noisy segmentation mask was replaced by the mask predicted by one of the CNNs if any one of them resulted in a lower loss; it was replaced by the average of the two CNN-predicted masks if both resulted in lower losses.

\end{itemize}

\subsubsection{Results}

The first row of Table \mbox{\ref{table:dwi_seg}} shows the DSC of the synthesized noisy segmentations, computed on the 65 images with manual segmentation. This can be regarded as an estimation of the accuracy of the training labels. The second row shows that strong label noise significantly affects the performance of the baseline CNN; the DSC achieved at test time always trails the DSC of the training labels. This is in disagreement with the results reported for handwritten digit recognition by \cite{rolnick2017}. As we reviewed above, \cite{rolnick2017} found that given sufficient training data with label accuracy slightly above random noise, classification accuracy of 90\% was achieved at test time. This difference is probably because our segmentation problem is more difficult and our training set is much smaller. Nonetheless, it is interesting to note that at the lowest label noise (noise level 1), the test DSC achieved by the baseline CNN (0.889) was higher than that achieved by the same model trained on the clean dataset (0.878), which consisted of approximately 40 times fewer images. For higher noise levels, training with MAE loss improved the classification results compared with the baseline CNN trained with the DSC loss. Dual CNN training with iterative label update performed consistently better than the baseline CNN and also performed much better than MAE loss on noise levels 1-5. For noise level 3-5, DSC achieved with this method was also higher than the DSC of the noisy labels that were used at the start of training. 

Table 5 shows more detailed performance measures on three different label noise levels. It shows the mean and standard deviation of the DSC and 95-percentile of the Hausdorff Distance (HD95), as well as the 5-percentile of the DSC (5\%  DSC) among the 65 test images. The results show that both MAE loss and Dual CNNs with iterative label update reduce the large segmentation errors, quantified with HD95, in the presence of strong label noise. There is also some improvement in 5\%  DSC, which is a measure of worst-case performance. Worst-case performance of the trained model is affected not only by the model accuracy, but also by the outlier data samples. Although the techniques reviewed in this paper and the methods used in our experiments are hoped to lead to better models that should perform better on average, the link to data outliers and difficult samples is less obvious. The great majority of the studies reviewed in this paper do not address the worst-case performance and data outliers, as those are essentially a different problem than label noise.

\begin{table*}[!htb]
\footnotesize
  \begin{center}
    \begin{tabular}{ L{4.5cm}  C{1.cm} C{1.2cm} C{1.1cm} C{1.1cm} C{1.1cm} C{1.1cm} C{1.1cm} C{1.1cm}  }
\hline \hline 
 & \scriptsize{Clean data} & \scriptsize{noise level 1 (Method 1)} & \scriptsize{noise level 2 (Method 2)} & \scriptsize{noise level 3 (Method 3)} & \scriptsize{noise level 4 (Method 2)} & \scriptsize{noise level 5 (Method 3)} &  \scriptsize{noise level 6 (Method 2)} & \scriptsize{noise level 7 (Method 2)}  \\ \hline 
Average DSC of the training labels & 1.000 & 0.949 & 0.924 & 0.854 & 0.807 & 0.790 & 0.777 & 0.742  \\ \hline
Baseline CNN & 0.878 & 0.889 & 0.862 & 0.846 & 0.755 & 0.730 & 0.736 & 0.724  \\
Baseline CNN trained with MAE loss & - & 0.881 & 0.864 & 0.840 &  0.780 & 0.741 & \textbf{0.778}  & \textbf{0.760}  \\
Dual CNNs with iterative label update & - & \textbf{0.906} & \textbf{0.895} & \textbf{0.886} &  \textbf{0.849} & \textbf{0.804} & 0.773 & 0.732  \\ \hline \hline 
\end{tabular}
  \end{center}
 \caption{\small{Comparison of different methods for fetal brain segmentation in DW-MR images in terms of DSC for different levels of label noise. The highest DSC scores have been highlighted in bold text for each noise level. Our dual CNNs with iterative label update generated the highest DSC scores at small and medium noise levels, whereas the CNN trained with MAE loss generated better results for high noise levels.}}
  \label{table:dwi_seg}
\end{table*}

\begin{table*}[!htb]
\scriptsize

    \begin{tabular}{ L{4.0cm}  C{1.1cm} C{0.9cm} C{1.3cm} C{1.1cm} C{0.9cm} C{1.3cm} C{1.1cm} C{0.9cm} C{1.3cm}  }
\hline \hline 
 & \multicolumn{3}{c}{noise level 1 (Method 1)} & \multicolumn{3}{c}{noise level 3 (Method 3)} & \multicolumn{3}{c}{noise level 5 (Method 3)}  \\ 
& DSC & 5\%  DSC & HD95 (mm) & DSC & 5\%  DSC & HD95 (mm) & DSC & 5\%  DSC & HD95 (mm)  \\ \hline
Baseline CNN &                           $ 0.89 \pm 0.06 $ & \boldmath$0.80$ & $ 5.9 \pm 2.6 $ & $ 0.85 \pm 0.08 $ & 0.73 & $ 6.8 \pm 2.6 $ & $ 0.73 \pm 0.10 $ & 0.60 & $ 8.0 \pm 4.9 $    \\
Baseline CNN trained with MAE loss  &    $ 0.88 \pm 0.07 $ & 0.79 & \boldmath$ 5.6 \pm 2.3 $ & $ 0.84 \pm 0.08 $ & 0.72 & $ 6.2 \pm 2.5 $ & $ 0.74 \pm 0.10 $ & 0.61 & $ 8.2 \pm 3.6 $    \\
Dual CNNs with iterative label update  & \boldmath$ 0.91 \pm 0.06 $ & 0.79 & $ 5.6 \pm 2.4 $ & \boldmath$ 0.89 \pm 0.08 $ & \boldmath$0.75$ & \boldmath$ 6.0 \pm 2.6 $ & \boldmath$ 0.80 \pm 0.11 $ & \boldmath$0.63$ & \boldmath$ 7.8 \pm 4.0 $    \\ \hline \hline 
\end{tabular}

  \label{table:dwi_seg2}

 \caption{\small{More detailed performance measures for fetal brain segmentation in DW-MRI. According to detailed analysis by three performance measures at different noise levels, our proposed dual CNNs with iterative label update outperformed both the baseline CNN and the baseline CNN trained with the MAE loss.}}

\end{table*}

\section{Discussion and Conclusions}
\label{conclusions}

Label noise is unavoidable in many medical image datasets. It can be caused by limited attention or expertise of the human annotator, subjective nature of labeling, or errors in computerized labeling systems. Since deep learning methods are increasingly used in medical image analysis, a proper understanding of the effects of label noise in training data and methods to manage those effects are essential. To help improve this understanding, this paper first presented a review of studies on label noise in machine learning and deep learning, followed by a review of studies on label noise in deep learning for medical image analysis; and second, investigated several existing and new methods and remedies to deal with different types of label noise in three different medical image datasets in detection, segmentation, and classification applications.

Our review of the literature shows that many studies have demonstrated negative effects of label noise in deep learning. Our review also shows that a diverse set of methods have been proposed and successfully applied to handle label noise in deep learning. Most of these methods have been developed for general computer vision and machine learning problems. Moreover, many of these methods have been evaluated on large-scale image classification datasets. Hence, reassessment of their performance for medical image analysis applications is warranted. Given the large variability in data size, label noise, and the nature of tasks that one may encounter in medical image analysis, it is likely that for each application one has to experiment with a number of methods to find the most suitable one. In spite of the need, our review of the literature shows that very few studies have directly addressed the issue of label noise in deep learning for medical image analysis. Therefore, motivated by the need, in a set of experiments reported in Section~\mbox{\ref{experiments}} we investigated and developed several existing strategies and new methods to reduce the negative impact of label noise in deep learning for different medical image analysis applications. Based on the results of our experiments and the literature, we make general recommendations as follow.

Label cleaning and pre-processing methods can be useful for most medical image analysis applications, but one has to be selective. Some methods in this category rely on prediction confidence for detecting incorrectly-labeled samples \mbox{\cite{northcutt2017}}, \mbox{\cite{ding2018}}. These methods can only be effective if the trained model has a well-calibrated prediction. Moreover, some methods in this category rely on matching a data sample or its feature vector with a set of data samples with clean labels \mbox{\cite{vo2015}} \mbox{\cite{lee2018}}. These methods may also have limited applicability in medical image analysis because data samples are larger in size and fewer in number, making the data matching more challenging due to the curse of dimensionality. On the other hand, methods such as that proposed by \mbox{\cite{veit2017}} could be useful in many detection and classification applications. Interestingly, in our experiments on brain lesion detection in Section \mbox{\ref{tsc_experiment}}, we achieved our best results by iterative label cleaning, indicating the great potential of these methods.

In the category of studies that suggest changing the network architecture, most methods introduce a noise layer or graphical model to learn the label noise statistics in parallel with model training. These methods are relatively easy to implement and evaluate. Yet, we are aware of only one study that has reported successfully employing such a method in medical image analysis \mbox{\cite{dgani2018}}. Nonetheless, we demonstrated the potential of these methods with our experiments in Section \mbox{\ref{pathology_experiment}}, where we obtained our best results with a method in this category involving estimation of the statistics of annotation error. Based on our results and those reported by studies in the machine learning and computer vision literature, we think methods in this category could be highly effective for classification and detection applications. Of particular utility to medical image analysis tasks are methods that enable estimation of labeling error of one or multiple annotators, such as the method of \mbox{\cite{tanno2019}} that we used in our experiments in Section \mbox{\ref{pathology_experiment}}.

Noise-robust loss functions have been mainly proposed as substitutes for cross-entropy loss for classification applications. Nonetheless, in addition to our pathology classification experiments (Section \mbox{\ref{pathology_experiment}}), such loss functions also proved to be useful in our fetal brain segmentation experiments (Section \mbox{\ref{segmentation_experiment}}).  In our experiments, we used MAE and iMAE loss functions, which are based on down-weighting data samples that are more likely to be incorrectly-labeled. More aggressive loss functions such as those proposed by \mbox{\cite{thulasidasan2019}} and \mbox{\cite{rusiecki2019}} could be more effective under very strong label noise. An advantage of these loss functions is that they are easy to quickly implement and evaluate.

Data re-weighting methods are also typically easy to implement. Several studies have already reported successful application of data re-weighting methods in medical image analysis ( \mbox{\cite{le2019}}, \mbox{\cite{xue2019}}, \mbox{\cite{zhu2019}}, \mbox{\cite{mirikharaji2019}}). In our own experiments, we implemented two variations of data re-weighting and found both of them to be effective. In experiments on lesion detection/segmentation (Section \mbox{\ref{tsc_experiment}}), we down-weighted data samples with high loss values, whereas in experiments on pathology classification (Section \mbox{\ref{pathology_experiment}}), where we had multiple labels for each data sample, we down-weighted high-loss labels. Such methods may be effective in many similar applications in medical image analysis.

Among the six categories of surveyed methods, those based on data and label consistency may be less applicable to medical image analysis tasks. Most of the proposed methods in this category are based on some measure of correlation or similarity between different data samples in the feature space. Due to the large dimensionality of the feature space in deep learning, these methods can suffer from the curse of dimensionality, as suggested by \mbox{\cite{chen2019}}. This problem can be more serious in medical image analysis due to the relatively large size of medical images and the relatively small number of samples.

Lastly, methods based on novel training procedures encompass a wide range of techniques that could be useful for almost all applications in medical image analysis. Given the diversity of methods in this category from the machine learning and computer vision literature, this seems to be an area with great potential for innovations and flexible application-specific solutions. Previous studies have developed and successfully evaluated such application-specific solutions for various medical image analysis tasks including segmentation (\mbox{\cite{min2018}}; \mbox{\cite{nie2018}}; \mbox{\cite{zhang2018b}}) and classification (\mbox{\cite{fries2019}}). Our proposed Dual CNNs with iterative label update, presented and tested in Section \mbox{\ref{segmentation_experiment}}, is a successful example of these methods for deep learning with noisy labels.

Deep learning for medical image analysis presents specific challenges that can be different from many computer vision and machine learning applications. These peculiarities may influence the choice of solutions for combating label noise as well. Our experiments in Section \ref{experiments} revealed some of these challenges. For example, an important characteristic of medical image datasets, in particular those carefully annotated by human experts, is their small size. The data size may have a complicated interplay with label noise. In our experiments on brain lesion segmentation in Section \ref{tsc_experiment}, a small (n=12) but carefully annotated training dataset resulted in a better model compared with a much larger (n=153) dataset with noisy annotations. By contrast, in our fetal brain segmentation experiment in Section \ref{segmentation_experiment}, more accurate models were trained using many images ($n \approx 2500$) with slightly noisy segmentations than using much fewer (n=65) images with manual segmentations. The interplay between the size and accuracy of the labeled training data also depends on the application. This warrants a reassessment of the optimal ways of obtaining labels from human experts or other means for each application. 

The data size may also influence the effectiveness of different strategies for handling label noise. For example, in several studies in computer vision that we reviewed in this paper, down-weighting or completely discarding data samples that were more likely to have incorrect labels proved to be an effective approach. This may be a less effective approach in medical imaging where datasets are relatively small. As shown in Table \ref{table:tsc_results}, for brain lesion segmentation we obtained better results by detecting and correcting missing annotations than by ignoring data samples with high loss values. For prostate digital pathology experiments in Section \ref{pathology_experiment}, where we had access to labels from six pathologists, ignoring high-loss labels proved effective. Nonetheless, on this dataset we achieved better performance by modeling annotator confusion rather than ignoring high-loss labels. For our fetal brain segmentation, too, we experimented with methods to down-weight or ignore segmentations that were more likely to be incorrect, but we did not achieve good results. Based on our experimental results and observations, it is better to improve the label accuracy or estimate the labeling error using techniques such as those we used in Sections \ref{tsc_experiment} and \ref{pathology_experiment} rather than to ignore data samples that are likely to have incorrect labels.

Another important consideration in medical image datasets is the subjective nature of annotation and the impact of inter-observer variability. If labels are obtained from a single expert, as in our experiments in Section \ref{tsc_experiment}, annotations may be systematically biased due to annotation habits or subjective opinion of a single annotator, risking generalizability when compared with the ``true label". The level of inter-observer variability depends significantly on factors such as the application, observer expertise, and attention \cite{gurari2015, lampert2016, donovan2013, nagpal2018}. Our experiments in Section \ref{pathology_experiment} targeted an application with known high inter-observer variability. Our results suggest that when labels from multiple experts are available, methods that model observer confusion as part of the training process generally perform better than methods that aggregate the labels in a separate step prior to training. Our results also showed significant gains due to using labels from multiple experts. 


Results of our experiments with brain lesion segmentation in Section \ref{tsc_experiment} and with digital pathology in Section \ref{pathology_experiment} share an important lesson. In both of these experiments, we achieved improved performance by modeling annotation error of the human expert(s). In Section \ref{tsc_experiment}, we observed that the annotator systematically missed smaller, fainter, and more isolated lesions. This is an expected behavior, and similar observations have been reported in previous studies \cite{robinson2016, quekel1999, kundel1976}. In our experiments, we exploited CNN prediction uncertainty, which enabled us to devise a novel and effective method to detect and fill in missing annotations in the training labels. Similar methods can be effective in training deep learning models for datasets with incomplete annotations, which are commonplace in medical image analysis. In \mbox{Section~\ref{pathology_experiment},} on the other hand, we exploited an approach originally proposed for general computer vision applications, and achieved very good performance. This method, which estimated the annotation error of individual experts in parallel with CNN training, proved to be more effective than several other methods including label fusion algorithms.

Our experiment on fetal brain segmentation in DW-MRI in Section \ref{segmentation_experiment} showed the potential value of computer-generated noisy labels. An interesting observation was that the baseline CNN achieved better results when trained with noisy segmentation masks transferred from the corresponding T2 images than when trained on 65 images that had been manually segmented. There are many situations in medical image analysis where such approximate annotations can be obtained at little or no cost from other images of the same subject, from matched subjects, or from an atlas. Our results demonstrate the potential utility of such annotations. Nonetheless, our results also showed that very inaccurate annotations led to poor training, indicating an important limitation of such labels.

In summary, in our experiments we investigated three common types of label noise in medical image datasets, and the relative effectiveness of several approaches to reduce the negative impact of label noise. The source, statistics, and strength of label noise in medical imaging is diverse; and our study shows that the effects of label noise should be carefully analyzed in training deep learning algorithms. This warrants further investigations and development of robust models and training algorithms.

\section{Acknowledgements}
\label{acknowledgements}

This study was supported in part by the National Institute of Biomedical Imaging and Bioengineering, and the National Institute of Neurological Disorders and Stroke of the National Institutes of Health (NIH) under Award Numbers R01EB018988, R01NS106030, and R01NS079788; and by a Technological Innovations in Neuroscience Award from the McKnight Foundation. The content is solely the responsibility of the authors and does not necessarily represent the official views of the NIH or the McKnight Foundation.

\bibliographystyle{IEEEtran}
\bibliography{davoodreferences}

\begin{thebibliography}{100}
\providecommand{\url}[1]{#1}
\csname url@samestyle\endcsname
\providecommand{\newblock}{\relax}
\providecommand{\bibinfo}[2]{#2}
\providecommand{\BIBentrySTDinterwordspacing}{\spaceskip=0pt\relax}
\providecommand{\BIBentryALTinterwordstretchfactor}{4}
\providecommand{\BIBentryALTinterwordspacing}{\spaceskip=\fontdimen2\font plus
\BIBentryALTinterwordstretchfactor\fontdimen3\font minus
  \fontdimen4\font\relax}
\providecommand{\BIBforeignlanguage}[2]{{%
\expandafter\ifx\csname l@#1\endcsname\relax
\typeout{** WARNING: IEEEtran.bst: No hyphenation pattern has been}%
\typeout{** loaded for the language `#1'. Using the pattern for}%
\typeout{** the default language instead.}%
\else
\language=\csname l@#1\endcsname
\fi
#2}}
\providecommand{\BIBdecl}{\relax}
\BIBdecl

\bibitem{ching2018}
T.~Ching, D.~S. Himmelstein, B.~K. Beaulieu-Jones, A.~A. Kalinin, B.~T. Do,
  G.~P. Way, E.~Ferrero, P.-M. Agapow, M.~Zietz, M.~M. Hoffman \emph{et~al.},
  ``Opportunities and obstacles for deep learning in biology and medicine,''
  \emph{Journal of The Royal Society Interface}, vol.~15, no. 141, p. 20170387,
  2018.

\bibitem{topol2019}
E.~J. Topol, ``High-performance medicine: the convergence of human and
  artificial intelligence,'' \emph{Nature medicine}, vol.~25, no.~1, p.~44,
  2019.

\bibitem{wang2017}
G.~Wang, M.~Kalra, and C.~G. Orton, ``Machine learning will transform radiology
  significantly within the next 5 years,'' \emph{Medical physics}, vol.~44,
  no.~6, pp. 2041--2044, 2017.

\bibitem{topol2019b}
E.~Topol, \emph{Deep medicine: how artificial intelligence can make healthcare
  human again}.\hskip 1em plus 0.5em minus 0.4em\relax Hachette UK, 2019.

\bibitem{prevedello2019}
L.~M. Prevedello, S.~S. Halabi, G.~Shih, C.~C. Wu, M.~D. Kohli, F.~H. Chokshi,
  B.~J. Erickson, J.~Kalpathy-Cramer, K.~P. Andriole, and A.~E. Flanders,
  ``Challenges related to artificial intelligence research in medical imaging
  and the importance of image analysis competitions,'' \emph{Radiology:
  Artificial Intelligence}, vol.~1, no.~1, p. e180031, 2019.

\bibitem{wang2018d}
G.~Wang, J.~C. Ye, K.~Mueller, and J.~A. Fessler, ``Image reconstruction is a
  new frontier of machine learning,'' \emph{IEEE transactions on medical
  imaging}, vol.~37, no.~6, pp. 1289--1296, 2018.

\bibitem{ronneberger2015}
O.~Ronneberger, P.~Fischer, and T.~Brox, ``U-net: Convolutional networks for
  biomedical image segmentation,'' in \emph{International Conference on Medical
  Image Computing and Computer-Assisted Intervention}, 2015, pp. 234--241.

\bibitem{haskins2019}
G.~Haskins, U.~Kruger, and P.~Yan, ``Deep learning in medical image
  registration: A survey,'' \emph{arXiv preprint arXiv:1903.02026}, 2019.

\bibitem{xu2019}
Y.~Xu, A.~Hosny, R.~Zeleznik, C.~Parmar, T.~Coroller, I.~Franco, R.~H. Mak, and
  H.~J. Aerts, ``Deep learning predicts lung cancer treatment response from
  serial medical imaging,'' \emph{Clinical Cancer Research}, vol.~25, no.~11,
  pp. 3266--3275, 2019.

\bibitem{mobadersany2018}
P.~Mobadersany, S.~Yousefi, M.~Amgad, D.~A. Gutman, J.~S. Barnholtz-Sloan,
  J.~E.~V. Vega, D.~J. Brat, and L.~A. Cooper, ``Predicting cancer outcomes
  from histology and genomics using convolutional networks,'' \emph{Proceedings
  of the National Academy of Sciences}, vol. 115, no.~13, pp. E2970--E2979,
  2018.

\bibitem{lecun2015}
Y.~LeCun, Y.~Bengio, and G.~Hinton, ``Deep learning,'' \emph{nature}, vol. 521,
  no. 7553, p. 436, 2015.

\bibitem{sun2017}
C.~Sun, A.~Shrivastava, S.~Singh, and A.~Gupta, ``Revisiting unreasonable
  effectiveness of data in deep learning era,'' in \emph{Proceedings of the
  IEEE international conference on computer vision}, 2017, pp. 843--852.

\bibitem{guo2016}
Y.~Guo, L.~Zhang, Y.~Hu, X.~He, and J.~Gao, ``Ms-celeb-1m: A dataset and
  benchmark for large-scale face recognition,'' in \emph{European Conference on
  Computer Vision}.\hskip 1em plus 0.5em minus 0.4em\relax Springer, 2016, pp.
  87--102.

\bibitem{deng2009}
J.~Deng, W.~Dong, R.~Socher, L.-J. Li, K.~Li, and L.~Fei-Fei, ``Imagenet: A
  large-scale hierarchical image database,'' in \emph{2009 IEEE conference on
  computer vision and pattern recognition}.\hskip 1em plus 0.5em minus
  0.4em\relax Ieee, 2009, pp. 248--255.

\bibitem{ipeirotis2010}
P.~G. Ipeirotis, F.~Provost, and J.~Wang, ``Quality management on amazon
  mechanical turk,'' in \emph{Proceedings of the ACM SIGKDD workshop on human
  computation}.\hskip 1em plus 0.5em minus 0.4em\relax ACM, 2010, pp. 64--67.

\bibitem{wang2018c}
F.~Wang, L.~Chen, C.~Li, S.~Huang, Y.~Chen, C.~Qian, and C.~Change~Loy, ``The
  devil of face recognition is in the noise,'' in \emph{Proceedings of the
  European Conference on Computer Vision (ECCV)}, 2018, pp. 765--780.

\bibitem{kuznetsova2018}
A.~Kuznetsova, H.~Rom, N.~Alldrin, J.~Uijlings, I.~Krasin, J.~Pont-Tuset,
  S.~Kamali, S.~Popov, M.~Malloci, T.~Duerig \emph{et~al.}, ``The open images
  dataset v4: Unified image classification, object detection, and visual
  relationship detection at scale,'' \emph{arXiv preprint arXiv:1811.00982},
  2018.

\bibitem{bridge2016}
P.~Bridge, A.~Fielding, P.~Rowntree, and A.~Pullar, ``Intraobserver
  variability: Should we worry?'' \emph{Journal of medical imaging and
  radiation sciences}, vol.~47, no.~3, pp. 217--220, 2016.

\bibitem{nir2018}
\BIBentryALTinterwordspacing
G.~Nir, S.~Hor, D.~Karimi, L.~Fazli, B.~F. Skinnider, P.~Tavassoli, D.~Turbin,
  C.~F. Villamil, G.~Wang, R.~S. Wilson, K.~A. Iczkowski, M.~S. Lucia, P.~C.
  Black, P.~Abolmaesumi, S.~L. Goldenberg, and S.~E. Salcudean, ``Automatic
  grading of prostate cancer in digitized histopathology images: Learning from
  multiple experts,'' \emph{Medical Image Analysis}, vol.~50, pp. 167 -- 180,
  2018. [Online]. Available:
  \url{http://www.sciencedirect.com/science/article/pii/S1361841518307497}
\BIBentrySTDinterwordspacing

\bibitem{gulshan2016}
V.~Gulshan, L.~Peng, M.~Coram, M.~C. Stumpe, D.~Wu, A.~Narayanaswamy,
  S.~Venugopalan, K.~Widner, T.~Madams, J.~Cuadros \emph{et~al.}, ``Development
  and validation of a deep learning algorithm for detection of diabetic
  retinopathy in retinal fundus photographs,'' \emph{Jama}, vol. 316, no.~22,
  pp. 2402--2410, 2016.

\bibitem{esteva2017}
A.~Esteva, B.~Kuprel, R.~A. Novoa, J.~Ko, S.~M. Swetter, H.~M. Blau, and
  S.~Thrun, ``Dermatologist-level classification of skin cancer with deep
  neural networks,'' \emph{Nature}, vol. 542, no. 7639, p. 115, 2017.

\bibitem{yan2018}
K.~Yan, X.~Wang, L.~Lu, and R.~M. Summers, ``Deeplesion: automated mining of
  large-scale lesion annotations and universal lesion detection with deep
  learning,'' \emph{Journal of Medical Imaging}, vol.~5, no.~3, p. 036501,
  2018.

\bibitem{irvin2019}
J.~Irvin, P.~Rajpurkar, M.~Ko, Y.~Yu, S.~Ciurea-Ilcus, C.~Chute, H.~Marklund,
  B.~Haghgoo, R.~Ball, K.~Shpanskaya \emph{et~al.}, ``Chexpert: A large chest
  radiograph dataset with uncertainty labels and expert comparison,''
  \emph{arXiv preprint arXiv:1901.07031}, 2019.

\bibitem{gurari2015}
D.~Gurari, D.~Theriault, M.~Sameki, B.~Isenberg, T.~A. Pham, A.~Purwada,
  P.~Solski, M.~Walker, C.~Zhang, J.~Y. Wong \emph{et~al.}, ``How to collect
  segmentations for biomedical images? a benchmark evaluating the performance
  of experts, crowdsourced non-experts, and algorithms,'' in \emph{2015 IEEE
  winter conference on applications of computer vision}.\hskip 1em plus 0.5em
  minus 0.4em\relax IEEE, 2015, pp. 1169--1176.

\bibitem{albarqouni2016}
S.~Albarqouni, C.~Baur, F.~Achilles, V.~Belagiannis, S.~Demirci, and N.~Navab,
  ``Aggnet: deep learning from crowds for mitosis detection in breast cancer
  histology images,'' \emph{IEEE transactions on medical imaging}, vol.~35,
  no.~5, pp. 1313--1321, 2016.

\bibitem{langlotz2019}
C.~P. Langlotz, B.~Allen, B.~J. Erickson, J.~Kalpathy-Cramer, K.~Bigelow, T.~S.
  Cook, A.~E. Flanders, M.~P. Lungren, D.~S. Mendelson, J.~D. Rudie
  \emph{et~al.}, ``A roadmap for foundational research on artificial
  intelligence in medical imaging: From the 2018 nih/rsna/acr/the academy
  workshop,'' \emph{Radiology}, vol. 291, no.~3, pp. 781--791, 2019.

\bibitem{ravi2016}
D.~Rav{\`\i}, C.~Wong, F.~Deligianni, M.~Berthelot, J.~Andreu-Perez, B.~Lo, and
  G.-Z. Yang, ``Deep learning for health informatics,'' \emph{IEEE journal of
  biomedical and health informatics}, vol.~21, no.~1, pp. 4--21, 2016.

\bibitem{wang2017a}
X.~Wang, Y.~Peng, L.~Lu, Z.~Lu, M.~Bagheri, and R.~M. Summers, ``Chestx-ray8:
  Hospital-scale chest x-ray database and benchmarks on weakly-supervised
  classification and localization of common thorax diseases,'' in
  \emph{Proceedings of the IEEE conference on computer vision and pattern
  recognition}, 2017, pp. 2097--2106.

\bibitem{cheplygina2019}
V.~Cheplygina, M.~de~Bruijne, and J.~P. Pluim, ``Not-so-supervised: a survey of
  semi-supervised, multi-instance, and transfer learning in medical image
  analysis,'' \emph{Medical image analysis}, vol.~54, pp. 280--296, 2019.

\bibitem{tajbakhsh2019}
N.~Tajbakhsh, L.~Jeyaseelan, Q.~Li, J.~Chiang, Z.~Wu, and X.~Ding, ``Embracing
  imperfect datasets: A review of deep learning solutions for medical image
  segmentation,'' \emph{arXiv preprint arXiv:1908.10454}, 2019.

\bibitem{frenay2013}
B.~Fr{\'e}nay and M.~Verleysen, ``Classification in the presence of label
  noise: a survey,'' \emph{IEEE transactions on neural networks and learning
  systems}, vol.~25, no.~5, pp. 845--869, 2013.

\bibitem{garcia2015}
S.~Garc{\'\i}a, J.~Luengo, and F.~Herrera, \emph{Data preprocessing in data
  mining}.\hskip 1em plus 0.5em minus 0.4em\relax Springer, 2015.

\bibitem{zhu2004b}
X.~Zhu and X.~Wu, ``Class noise vs. attribute noise: A quantitative study,''
  \emph{Artificial intelligence review}, vol.~22, no.~3, pp. 177--210, 2004.

\bibitem{quinlan1986}
J.~R. Quinlan, ``Induction of decision trees,'' \emph{Machine learning},
  vol.~1, no.~1, pp. 81--106, 1986.

\bibitem{nettleton2010}
D.~F. Nettleton, A.~Orriols-Puig, and A.~Fornells, ``A study of the effect of
  different types of noise on the precision of supervised learning
  techniques,'' \emph{Artificial intelligence review}, vol.~33, no.~4, pp.
  275--306, 2010.

\bibitem{folleco2008}
A.~Folleco, T.~M. Khoshgoftaar, J.~Van~Hulse, and L.~Bullard, ``Identifying
  learners robust to low quality data,'' in \emph{2008 IEEE International
  Conference on Information Reuse and Integration}.\hskip 1em plus 0.5em minus
  0.4em\relax IEEE, 2008, pp. 190--195.

\bibitem{abellan2010}
J.~Abell{\'a}n and A.~R. Masegosa, ``Bagging decision trees on data sets with
  classification noise,'' in \emph{International Symposium on Foundations of
  Information and Knowledge Systems}.\hskip 1em plus 0.5em minus 0.4em\relax
  Springer, 2010, pp. 248--265.

\bibitem{mcdonald2003}
R.~A. McDonald, D.~J. Hand, and I.~A. Eckley, ``An empirical comparison of
  three boosting algorithms on real data sets with artificial class noise,'' in
  \emph{International Workshop on Multiple Classifier Systems}.\hskip 1em plus
  0.5em minus 0.4em\relax Springer, 2003, pp. 35--44.

\bibitem{long2010b}
P.~M. Long and R.~A. Servedio, ``Random classification noise defeats all convex
  potential boosters,'' \emph{Machine learning}, vol.~78, no.~3, pp. 287--304,
  2010.

\bibitem{dietterich2000}
T.~G. Dietterich, ``An experimental comparison of three methods for
  constructing ensembles of decision trees: Bagging, boosting, and
  randomization,'' \emph{Machine learning}, vol.~40, no.~2, pp. 139--157, 2000.

\bibitem{manwani2013}
N.~Manwani and P.~Sastry, ``Noise tolerance under risk minimization,''
  \emph{IEEE transactions on cybernetics}, vol.~43, no.~3, pp. 1146--1151,
  2013.

\bibitem{patrini2016}
G.~Patrini, F.~Nielsen, R.~Nock, and M.~Carioni, ``Loss factorization, weakly
  supervised learning and label noise robustness,'' in \emph{International
  conference on machine learning}, 2016, pp. 708--717.

\bibitem{van2015b}
B.~Van~Rooyen, A.~Menon, and R.~C. Williamson, ``Learning with symmetric label
  noise: The importance of being unhinged,'' in \emph{Advances in Neural
  Information Processing Systems}, 2015, pp. 10--18.

\bibitem{liu2015}
T.~Liu and D.~Tao, ``Classification with noisy labels by importance
  reweighting,'' \emph{IEEE Transactions on pattern analysis and machine
  intelligence}, vol.~38, no.~3, pp. 447--461, 2015.

\bibitem{natarajan2013}
N.~Natarajan, I.~S. Dhillon, P.~K. Ravikumar, and A.~Tewari, ``Learning with
  noisy labels,'' in \emph{Advances in neural information processing systems},
  2013, pp. 1196--1204.

\bibitem{segata2009}
N.~Segata, E.~Blanzieri, and P.~Cunningham, ``A scalable noise reduction
  technique for large case-based systems,'' in \emph{International Conference
  on Case-Based Reasoning}.\hskip 1em plus 0.5em minus 0.4em\relax Springer,
  2009, pp. 328--342.

\bibitem{brodley1996}
C.~E. Brodley, M.~A. Friedl \emph{et~al.}, ``Identifying and eliminating
  mislabeled training instances,'' in \emph{Proceedings of the National
  Conference on Artificial Intelligence}, 1996, pp. 799--805.

\bibitem{sluban2010}
B.~Sluban, D.~Gamberger, and N.~Lavra, ``Advances in class noise detection,''
  in \emph{Proceedings of the 2010 conference on ECAI 2010: 19th European
  Conference on Artificial Intelligence}.\hskip 1em plus 0.5em minus
  0.4em\relax IOS Press, 2010, pp. 1105--1106.

\bibitem{wilson1997}
D.~R. Wilson and T.~R. Martinez, ``Instance pruning techniques,'' in
  \emph{ICML}, vol.~97, no. 1997, 1997, pp. 400--411.

\bibitem{wilson2000}
------, ``Reduction techniques for instance-based learning algorithms,''
  \emph{Machine learning}, vol.~38, no.~3, pp. 257--286, 2000.

\bibitem{zhang2009}
C.~Zhang, C.~Wu, E.~Blanzieri, Y.~Zhou, Y.~Wang, W.~Du, and Y.~Liang, ``Methods
  for labeling error detection in microarrays based on the effect of data
  perturbation on the regression model,'' \emph{Bioinformatics}, vol.~25,
  no.~20, pp. 2708--2714, 2009.

\bibitem{malossini2006}
A.~Malossini, E.~Blanzieri, and R.~T. Ng, ``Detecting potential labeling errors
  in microarrays by data perturbation,'' \emph{Bioinformatics}, vol.~22,
  no.~17, pp. 2114--2121, 2006.

\bibitem{gamberger2000}
D.~Gamberger, N.~Lavrac, and S.~Dzeroski, ``Noise detection and elimination in
  data preprocessing: experiments in medical domains,'' \emph{Applied
  Artificial Intelligence}, vol.~14, no.~2, pp. 205--223, 2000.

\bibitem{sun2007}
J.-w. Sun, F.-y. Zhao, C.-j. Wang, and S.-f. Chen, ``Identifying and correcting
  mislabeled training instances,'' in \emph{Future generation communication and
  networking (FGCN 2007)}, vol.~1.\hskip 1em plus 0.5em minus 0.4em\relax IEEE,
  2007, pp. 244--250.

\bibitem{khardon2007}
R.~Khardon and G.~Wachman, ``Noise tolerant variants of the perceptron
  algorithm,'' \emph{Journal of Machine Learning Research}, vol.~8, no. Feb,
  pp. 227--248, 2007.

\bibitem{lin2004}
C.-f. Lin \emph{et~al.}, ``Training algorithms for fuzzy support vector
  machines with noisy data,'' \emph{Pattern recognition letters}, vol.~25,
  no.~14, pp. 1647--1656, 2004.

\bibitem{kaster2010}
F.~O. Kaster, B.~H. Menze, M.-A. Weber, and F.~A. Hamprecht, ``Comparative
  validation of graphical models for learning tumor segmentations from noisy
  manual annotations,'' in \emph{International MICCAI Workshop on Medical
  Computer Vision}.\hskip 1em plus 0.5em minus 0.4em\relax Springer, 2010, pp.
  74--85.

\bibitem{kim2006}
H.-C. Kim and Z.~Ghahramani, ``Bayesian gaussian process classification with
  the em-ep algorithm,'' \emph{IEEE Transactions on Pattern Analysis and
  Machine Intelligence}, vol.~28, no.~12, pp. 1948--1959, 2006.

\bibitem{zhang2016}
C.~Zhang, S.~Bengio, M.~Hardt, B.~Recht, and O.~Vinyals, ``Understanding deep
  learning requires rethinking generalization,'' \emph{arXiv preprint
  arXiv:1611.03530}, 2016.

\bibitem{arpit2017}
D.~Arpit, S.~Jastrzebski, N.~Ballas, D.~Krueger, E.~Bengio, M.~S. Kanwal,
  T.~Maharaj, A.~Fischer, A.~Courville, Y.~Bengio \emph{et~al.}, ``A closer
  look at memorization in deep networks,'' in \emph{Proceedings of the 34th
  International Conference on Machine Learning-Volume 70}, 2017, pp. 233--242.

\bibitem{rolnick2017}
D.~Rolnick, A.~Veit, S.~Belongie, and N.~Shavit, ``Deep learning is robust to
  massive label noise,'' \emph{arXiv preprint arXiv:1705.10694}, 2017.

\bibitem{chen2019b}
P.~Chen, B.~Liao, G.~Chen, and S.~Zhang, ``Understanding and utilizing deep
  neural networks trained with noisy labels,'' \emph{arXiv preprint
  arXiv:1905.05040}, 2019.

\bibitem{ma2018}
X.~Ma, Y.~Wang, M.~E. Houle, S.~Zhou, S.~M. Erfani, S.-T. Xia, S.~Wijewickrema,
  and J.~Bailey, ``Dimensionality-driven learning with noisy labels,''
  \emph{arXiv preprint arXiv:1806.02612}, 2018.

\bibitem{houle2017}
M.~E. Houle, ``Local intrinsic dimensionality i: an extreme-value-theoretic
  foundation for similarity applications,'' in \emph{International Conference
  on Similarity Search and Applications}.\hskip 1em plus 0.5em minus
  0.4em\relax Springer, 2017, pp. 64--79.

\bibitem{drory2018}
A.~Drory, S.~Avidan, and R.~Giryes, ``On the resistance of neural nets to label
  noise,'' \emph{arXiv preprint arXiv:1803.11410}, 2018.

\bibitem{martin2017}
C.~H. Martin and M.~W. Mahoney, ``Rethinking generalization requires revisiting
  old ideas: statistical mechanics approaches and complex learning behavior,''
  \emph{arXiv preprint arXiv:1710.09553}, 2017.

\bibitem{yu2017a}
X.~Yu, T.~Liu, M.~Gong, K.~Zhang, and D.~Tao, ``Transfer learning with label
  noise,'' \emph{arXiv preprint arXiv:1707.09724}, 2017.

\bibitem{moosavi2017}
S.-M. Moosavi-Dezfooli, A.~Fawzi, O.~Fawzi, and P.~Frossard, ``Universal
  adversarial perturbations,'' in \emph{Proceedings of the IEEE conference on
  computer vision and pattern recognition}, 2017, pp. 1765--1773.

\bibitem{speth2019}
J.~Speth and E.~M. Hand, ``Automated label noise identification for facial
  attribute recognition,'' in \emph{Proceedings of the IEEE Conference on
  Computer Vision and Pattern Recognition Workshops}, 2019, pp. 25--28.

\bibitem{ostyakov2018}
P.~Ostyakov, E.~Logacheva, R.~Suvorov, V.~Aliev, G.~Sterkin, O.~Khomenko, and
  S.~I. Nikolenko, ``Label denoising with large ensembles of heterogeneous
  neural networks,'' in \emph{Proceedings of the European Conference on
  Computer Vision (ECCV)}, 2018, pp. 0--0.

\bibitem{pham2019}
H.~H. Pham, T.~T. Le, D.~Q. Tran, D.~T. Ngo, and H.~Q. Nguyen, ``Interpreting
  chest x-rays via cnns that exploit disease dependencies and uncertainty
  labels,'' \emph{arXiv preprint arXiv:1911.06475}, 2019.

\bibitem{lee2018}
K.-H. Lee, X.~He, L.~Zhang, and L.~Yang, ``Cleannet: Transfer learning for
  scalable image classifier training with label noise,'' in \emph{Proceedings
  of the IEEE Conference on Computer Vision and Pattern Recognition}, 2018, pp.
  5447--5456.

\bibitem{northcutt2017}
C.~G. Northcutt, T.~Wu, and I.~L. Chuang, ``Learning with confident examples:
  Rank pruning for robust classification with noisy labels,'' \emph{arXiv
  preprint arXiv:1705.01936}, 2017.

\bibitem{veit2017}
A.~Veit, N.~Alldrin, G.~Chechik, I.~Krasin, A.~Gupta, and S.~Belongie,
  ``Learning from noisy large-scale datasets with minimal supervision,'' in
  \emph{Proceedings of the IEEE Conference on Computer Vision and Pattern
  Recognition}, 2017, pp. 839--847.

\bibitem{gao2017}
B.-B. Gao, C.~Xing, C.-W. Xie, J.~Wu, and X.~Geng, ``Deep label distribution
  learning with label ambiguity,'' \emph{IEEE Transactions on Image
  Processing}, vol.~26, no.~6, pp. 2825--2838, 2017.

\bibitem{sukhbaatar2014b}
S.~Sukhbaatar and R.~Fergus, ``Learning from noisy labels with deep neural
  networks,'' \emph{arXiv preprint arXiv:1406.2080}, vol.~2, no.~3, p.~4, 2014.

\bibitem{dgani2018}
Y.~Dgani, H.~Greenspan, and J.~Goldberger, ``Training a neural network based on
  unreliable human annotation of medical images,'' in \emph{2018 IEEE 15th
  International Symposium on Biomedical Imaging (ISBI 2018)}.\hskip 1em plus
  0.5em minus 0.4em\relax IEEE, 2018, pp. 39--42.

\bibitem{vahdat2017}
A.~Vahdat, ``Toward robustness against label noise in training deep
  discriminative neural networks,'' in \emph{Advances in Neural Information
  Processing Systems}, 2017, pp. 5596--5605.

\bibitem{yao2018}
J.~Yao, J.~Wang, I.~W. Tsang, Y.~Zhang, J.~Sun, C.~Zhang, and R.~Zhang, ``Deep
  learning from noisy image labels with quality embedding,'' \emph{IEEE
  Transactions on Image Processing}, vol.~28, no.~4, pp. 1909--1922, 2018.

\bibitem{ghosh2017}
A.~Ghosh, H.~Kumar, and P.~Sastry, ``Robust loss functions under label noise
  for deep neural networks,'' in \emph{Thirty-First AAAI Conference on
  Artificial Intelligence}, 2017.

\bibitem{matuszewski2018}
D.~J. Matuszewski and I.-M. Sintorn, ``Minimal annotation training for
  segmentation of microscopy images,'' in \emph{2018 IEEE 15th International
  Symposium on Biomedical Imaging (ISBI 2018)}.\hskip 1em plus 0.5em minus
  0.4em\relax IEEE, 2018, pp. 387--390.

\bibitem{zhang2018}
Z.~Zhang and M.~Sabuncu, ``Generalized cross entropy loss for training deep
  neural networks with noisy labels,'' in \emph{Advances in Neural Information
  Processing Systems}, 2018, pp. 8778--8788.

\bibitem{wang2019c}
X.~Wang, E.~Kodirov, Y.~Hua, and N.~M. Robertson, ``Improving mae against cce
  under label noise,'' \emph{arXiv preprint arXiv:1903.12141}, 2019.

\bibitem{rusiecki2019}
A.~Rusiecki, ``Trimmed robust loss function for training deep neural networks
  with label noise,'' in \emph{International Conference on Artificial
  Intelligence and Soft Computing}.\hskip 1em plus 0.5em minus 0.4em\relax
  Springer, 2019, pp. 215--222.

\bibitem{boughorbel2018}
S.~Boughorbel, F.~Jarray, N.~Venugopal, and H.~Elhadi, ``Alternating loss
  correction for preterm-birth prediction from ehr data with noisy labels,''
  \emph{arXiv preprint arXiv:1811.09782}, 2018.

\bibitem{hendrycks2018}
D.~Hendrycks, M.~Mazeika, D.~Wilson, and K.~Gimpel, ``Using trusted data to
  train deep networks on labels corrupted by severe noise,'' in \emph{Advances
  in Neural Information Processing Systems}, 2018, pp. 10\,456--10\,465.

\bibitem{ren2018}
M.~Ren, W.~Zeng, B.~Yang, and R.~Urtasun, ``Learning to reweight examples for
  robust deep learning,'' \emph{arXiv preprint arXiv:1803.09050}, 2018.

\bibitem{le2019}
H.~Le, D.~Samaras, T.~Kurc, R.~Gupta, K.~Shroyer, and J.~Saltz, ``Pancreatic
  cancer detection in whole slide images using noisy label annotations,'' in
  \emph{Medical Image Computing and Computer Assisted Intervention -- MICCAI
  2019}.\hskip 1em plus 0.5em minus 0.4em\relax Springer International
  Publishing, 2019.

\bibitem{xue2019}
C.~Xue, Q.~Dou, X.~Shi, H.~Chen, and P.~A. Heng, ``Robust learning at noisy
  labeled medical images: Applied to skin lesion classification,'' \emph{arXiv
  preprint arXiv:1901.07759}, 2019.

\bibitem{zhu2019}
H.~Zhu, J.~Shi, and J.~Wu, ``Pick-and-learn: Automatic quality evaluation for
  noisy-labeled image segmentation,'' \emph{arXiv preprint arXiv:1907.11835},
  2019.

\bibitem{mirikharaji2019}
Z.~Mirikharaji, Y.~Yan, and G.~Hamarneh, ``Learning to segment skin lesions
  from noisy annotations,'' \emph{arXiv preprint arXiv:1906.03815}, 2019.

\bibitem{shu2019}
J.~Shu, Q.~Xie, L.~Yi, Q.~Zhao, S.~Zhou, Z.~Xu, and D.~Meng, ``Meta-weight-net:
  Learning an explicit mapping for sample weighting,'' \emph{arXiv preprint
  arXiv:1902.07379}, 2019.

\bibitem{khetan2017}
A.~Khetan, Z.~C. Lipton, and A.~Anandkumar, ``Learning from noisy
  singly-labeled data,'' \emph{arXiv preprint arXiv:1712.04577}, 2017.

\bibitem{tanno2019}
R.~Tanno, A.~Saeedi, S.~Sankaranarayanan, D.~C. Alexander, and N.~Silberman,
  ``Learning from noisy labels by regularized estimation of annotator
  confusion,'' \emph{arXiv preprint arXiv:1902.03680}, 2019.

\bibitem{shen2019}
Y.~Shen and S.~Sanghavi, ``Learning with bad training data via iterative
  trimmed loss minimization,'' in \emph{International Conference on Machine
  Learning}, 2019, pp. 5739--5748.

\bibitem{lee2019}
K.~Lee, S.~Yun, K.~Lee, H.~Lee, B.~Li, and J.~Shin, ``Robust inference via
  generative classifiers for handling noisy labels,'' \emph{arXiv preprint
  arXiv:1901.11300}, 2019.

\bibitem{yu2019b}
L.~Yu, S.~Wang, X.~Li, C.-W. Fu, and P.-A. Heng, ``Uncertainty-aware
  self-ensembling model for semi-supervised 3d left atrium segmentation,''
  \emph{arXiv preprint arXiv:1907.07034}, 2019.

\bibitem{zhang2019}
W.~Zhang, Y.~Wang, and Y.~Qiao, ``Metacleaner: Learning to hallucinate clean
  representations for noisy-labeled visual recognition,'' in \emph{Proceedings
  of the IEEE Conference on Computer Vision and Pattern Recognition}, 2019, pp.
  7373--7382.

\bibitem{azadi2015}
S.~Azadi, J.~Feng, S.~Jegelka, and T.~Darrell, ``Auxiliary image regularization
  for deep cnns with noisy labels,'' \emph{arXiv preprint arXiv:1511.07069},
  2015.

\bibitem{wang2018b}
Y.~Wang, W.~Liu, X.~Ma, J.~Bailey, H.~Zha, L.~Song, and S.-T. Xia, ``Iterative
  learning with open-set noisy labels,'' in \emph{Proceedings of the IEEE
  Conference on Computer Vision and Pattern Recognition}, 2018, pp. 8688--8696.

\bibitem{reed2014}
S.~Reed, H.~Lee, D.~Anguelov, C.~Szegedy, D.~Erhan, and A.~Rabinovich,
  ``Training deep neural networks on noisy labels with bootstrapping,''
  \emph{arXiv preprint arXiv:1412.6596}, 2014.

\bibitem{zhong2019}
Y.~Zhong, W.~Deng, M.~Wang, J.~Hu, J.~Peng, X.~Tao, and Y.~Huang,
  ``Unequal-training for deep face recognition with long-tailed noisy data,''
  in \emph{Proceedings of the IEEE Conference on Computer Vision and Pattern
  Recognition}, 2019, pp. 7812--7821.

\bibitem{min2018}
S.~Min, X.~Chen, Z.-J. Zha, F.~Wu, and Y.~Zhang, ``A two-stream mutual
  attention network for semi-supervised biomedical segmentation with noisy
  labels,'' \emph{arXiv preprint arXiv:1807.11719}, 2018.

\bibitem{nie2018}
D.~Nie, Y.~Gao, L.~Wang, and D.~Shen, ``Asdnet: Attention based semi-supervised
  deep networks for medical image segmentation,'' in \emph{International
  Conference on Medical Image Computing and Computer-Assisted
  Intervention}.\hskip 1em plus 0.5em minus 0.4em\relax Springer, 2018, pp.
  370--378.

\bibitem{zhang2018b}
L.~Zhang, V.~Gopalakrishnan, L.~Lu, R.~M. Summers, J.~Moss, and J.~Yao,
  ``Self-learning to detect and segment cysts in lung ct images without manual
  annotation,'' in \emph{2018 IEEE 15th International Symposium on Biomedical
  Imaging (ISBI 2018)}.\hskip 1em plus 0.5em minus 0.4em\relax IEEE, 2018, pp.
  1100--1103.

\bibitem{fries2019}
J.~A. Fries, P.~Varma, V.~S. Chen, K.~Xiao, H.~Tejeda, P.~Saha, J.~Dunnmon,
  H.~Chubb, S.~Maskatia, M.~Fiterau \emph{et~al.}, ``Weakly supervised
  classification of aortic valve malformations using unlabeled cardiac mri
  sequences,'' \emph{BioRxiv}, p. 339630, 2019.

\bibitem{jiang2017}
L.~Jiang, Z.~Zhou, T.~Leung, L.-J. Li, and L.~Fei-Fei, ``Mentornet: Learning
  data-driven curriculum for very deep neural networks on corrupted labels,''
  \emph{arXiv preprint arXiv:1712.05055}, 2017.

\bibitem{han2018}
B.~Han, Q.~Yao, X.~Yu, G.~Niu, M.~Xu, W.~Hu, I.~Tsang, and M.~Sugiyama,
  ``Co-teaching: Robust training of deep neural networks with extremely noisy
  labels,'' in \emph{Advances in Neural Information Processing Systems}, 2018,
  pp. 8527--8537.

\bibitem{zhang2017}
H.~Zhang, M.~Cisse, Y.~N. Dauphin, and D.~Lopez-Paz, ``mixup: Beyond empirical
  risk minimization,'' \emph{arXiv preprint arXiv:1710.09412}, 2017.

\bibitem{acuna2019}
D.~Acuna, A.~Kar, and S.~Fidler, ``Devil is in the edges: Learning semantic
  boundaries from noisy annotations,'' in \emph{Proceedings of the IEEE
  Conference on Computer Vision and Pattern Recognition}, 2019, pp.
  11\,075--11\,083.

\bibitem{yu2018}
Z.~Yu, W.~Liu, Y.~Zou, C.~Feng, S.~Ramalingam, B.~Vijaya~Kumar, and J.~Kautz,
  ``Simultaneous edge alignment and learning,'' in \emph{Proceedings of the
  European Conference on Computer Vision (ECCV)}, 2018, pp. 388--404.

\bibitem{vo2015}
P.~D. Vo, A.~Ginsca, H.~Le~Borgne, and A.~Popescu, ``Effective training of
  convolutional networks using noisy web images,'' in \emph{2015 13th
  International Workshop on Content-Based Multimedia Indexing (CBMI)}.\hskip
  1em plus 0.5em minus 0.4em\relax IEEE, 2015, pp. 1--6.

\bibitem{han2019}
J.~Han, P.~Luo, and X.~Wang, ``Deep self-learning from noisy labels,''
  \emph{arXiv preprint arXiv:1908.02160}, 2019.

\bibitem{guo2017}
C.~Guo, G.~Pleiss, Y.~Sun, and K.~Q. Weinberger, ``On calibration of modern
  neural networks,'' \emph{arXiv preprint arXiv:1706.04599}, 2017.

\bibitem{lakshminarayanan2017}
B.~Lakshminarayanan, A.~Pritzel, and C.~Blundell, ``Simple and scalable
  predictive uncertainty estimation using deep ensembles,'' in \emph{Advances
  in Neural Information Processing Systems}, 2017, pp. 6402--6413.

\bibitem{gal2015}
Y.~Gal and Z.~Ghahramani, ``Bayesian convolutional neural networks with
  bernoulli approximate variational inference,'' \emph{arXiv preprint
  arXiv:1506.02158}, 2015.

\bibitem{kendall2017}
A.~Kendall and Y.~Gal, ``What uncertainties do we need in bayesian deep
  learning for computer vision?'' in \emph{Advances in neural information
  processing systems}, 2017, pp. 5574--5584.

\bibitem{pawlowski2017}
N.~Pawlowski, A.~Brock, M.~C. Lee, M.~Rajchl, and B.~Glocker, ``Implicit weight
  uncertainty in neural networks,'' \emph{arXiv preprint arXiv:1711.01297},
  2017.

\bibitem{ding2018}
Y.~Ding, L.~Wang, D.~Fan, and B.~Gong, ``A semi-supervised two-stage approach
  to learning from noisy labels,'' in \emph{2018 IEEE Winter Conference on
  Applications of Computer Vision (WACV)}.\hskip 1em plus 0.5em minus
  0.4em\relax IEEE, 2018, pp. 1215--1224.

\bibitem{kohler2019}
J.~M. K{\"o}hler, M.~Autenrieth, and W.~H. Beluch, ``Uncertainty based
  detection and relabeling of noisy image labels,'' in \emph{Proceedings of the
  IEEE Conference on Computer Vision and Pattern Recognition Workshops}, 2019,
  pp. 33--37.

\bibitem{szegedy2016}
C.~Szegedy, V.~Vanhoucke, S.~Ioffe, J.~Shlens, and Z.~Wojna, ``Rethinking the
  inception architecture for computer vision,'' in \emph{Proceedings of the
  IEEE conference on computer vision and pattern recognition}, 2016, pp.
  2818--2826.

\bibitem{muller2019}
R.~M{\"u}ller, S.~Kornblith, and G.~Hinton, ``When does label smoothing help?''
  \emph{arXiv preprint arXiv:1906.02629}, 2019.

\bibitem{yi2019}
K.~Yi and J.~Wu, ``Probabilistic end-to-end noise correction for learning with
  noisy labels,'' \emph{arXiv preprint arXiv:1903.07788}, 2019.

\bibitem{ratner2016}
A.~J. Ratner, C.~M. De~Sa, S.~Wu, D.~Selsam, and C.~R{\'e}, ``Data programming:
  Creating large training sets, quickly,'' in \emph{Advances in neural
  information processing systems}, 2016, pp. 3567--3575.

\bibitem{zhou2017b}
H.~Zhou, J.~Sun, Y.~Yacoob, and D.~W. Jacobs, ``Label denoising adversarial
  network (ldan) for inverse lighting of face images,'' \emph{arXiv preprint
  arXiv:1709.01993}, 2017.

\bibitem{chiaroni2019}
F.~Chiaroni, M.~Rahal, N.~Hueber, and F.~Dufaux, ``Hallucinating a cleanly
  labeled augmented dataset from a noisy labeled dataset using gans,'' 2019.

\bibitem{sukhbaatar2014}
S.~Sukhbaatar, J.~Bruna, M.~Paluri, L.~Bourdev, and R.~Fergus, ``Training
  convolutional networks with noisy labels,'' \emph{arXiv preprint
  arXiv:1406.2080}, 2014.

\bibitem{thekumparampil2018}
K.~K. Thekumparampil, A.~Khetan, Z.~Lin, and S.~Oh, ``Robustness of conditional
  gans to noisy labels,'' in \emph{Advances in Neural Information Processing
  Systems}, 2018, pp. 10\,271--10\,282.

\bibitem{goldberger2016}
J.~Goldberger and E.~Ben-Reuven, ``Training deep neural-networks using a noise
  adaptation layer,'' 2016.

\bibitem{bekker2016}
A.~J. Bekker and J.~Goldberger, ``Training deep neural-networks based on
  unreliable labels,'' in \emph{2016 IEEE International Conference on
  Acoustics, Speech and Signal Processing (ICASSP)}.\hskip 1em plus 0.5em minus
  0.4em\relax IEEE, 2016, pp. 2682--2686.

\bibitem{jindal2016}
I.~Jindal, M.~Nokleby, and X.~Chen, ``Learning deep networks from noisy labels
  with dropout regularization,'' in \emph{2016 IEEE 16th International
  Conference on Data Mining (ICDM)}.\hskip 1em plus 0.5em minus 0.4em\relax
  IEEE, 2016, pp. 967--972.

\bibitem{xiao2015}
T.~Xiao, T.~Xia, Y.~Yang, C.~Huang, and X.~Wang, ``Learning from massive noisy
  labeled data for image classification,'' in \emph{Proceedings of the IEEE
  conference on computer vision and pattern recognition}, 2015, pp. 2691--2699.

\bibitem{misra2016}
I.~Misra, C.~Lawrence~Zitnick, M.~Mitchell, and R.~Girshick, ``Seeing through
  the human reporting bias: Visual classifiers from noisy human-centric
  labels,'' in \emph{Proceedings of the IEEE Conference on Computer Vision and
  Pattern Recognition}, 2016, pp. 2930--2939.

\bibitem{izadinia2015}
H.~Izadinia, B.~C. Russell, A.~Farhadi, M.~D. Hoffman, and A.~Hertzmann, ``Deep
  classifiers from image tags in the wild,'' in \emph{Proceedings of the 2015
  Workshop on Community-Organized Multimodal Mining: Opportunities for Novel
  Solutions}.\hskip 1em plus 0.5em minus 0.4em\relax ACM, 2015, pp. 13--18.

\bibitem{thulasidasan2019}
S.~Thulasidasan, T.~Bhattacharya, J.~Bilmes, G.~Chennupati, and J.~Mohd-Yusof,
  ``Combating label noise in deep learning using abstention,'' \emph{arXiv
  preprint arXiv:1905.10964}, 2019.

\bibitem{freund1999}
Y.~Freund, R.~Schapire, and N.~Abe, ``A short introduction to boosting,''
  \emph{Journal-Japanese Society For Artificial Intelligence}, vol.~14, no.
  771-780, p. 1612, 1999.

\bibitem{shrivastava2016}
A.~Shrivastava, A.~Gupta, and R.~Girshick, ``Training region-based object
  detectors with online hard example mining,'' in \emph{Proceedings of the IEEE
  conference on computer vision and pattern recognition}, 2016, pp. 761--769.

\bibitem{lin2017}
T.-Y. Lin, P.~Goyal, R.~Girshick, K.~He, and P.~Doll{\'a}r, ``Focal loss for
  dense object detection,'' in \emph{Proceedings of the IEEE international
  conference on computer vision}, 2017, pp. 2980--2988.

\bibitem{patrini2017}
G.~Patrini, A.~Rozza, A.~Krishna~Menon, R.~Nock, and L.~Qu, ``Making deep
  neural networks robust to label noise: A loss correction approach,'' in
  \emph{Proceedings of the IEEE Conference on Computer Vision and Pattern
  Recognition}, 2017, pp. 1944--1952.

\bibitem{reid2010}
M.~D. Reid and R.~C. Williamson, ``Composite binary losses,'' \emph{Journal of
  Machine Learning Research}, vol.~11, no. Sep, pp. 2387--2422, 2010.

\bibitem{mnih2012}
V.~Mnih and G.~E. Hinton, ``Learning to label aerial images from noisy data,''
  in \emph{Proceedings of the 29th International conference on machine learning
  (ICML-12)}, 2012, pp. 567--574.

\bibitem{wang2019b}
X.~Wang, Y.~Hua, E.~Kodirov, and N.~Robertson, ``Emphasis regularisation by
  gradient rescaling for training deep neural networks with noisy labels,''
  \emph{arXiv preprint arXiv:1905.11233}, 2019.

\bibitem{han2018b}
B.~Han, G.~Niu, J.~Yao, X.~Yu, M.~Xu, I.~Tsang, and M.~Sugiyama, ``Pumpout: A
  meta approach for robustly training deep neural networks with noisy labels,''
  \emph{arXiv preprint arXiv:1809.11008}, 2018.

\bibitem{warfield2004}
S.~K. Warfield, K.~H. Zou, and W.~M. Wells, ``Simultaneous truth and
  performance level estimation (staple): an algorithm for the validation of
  image segmentation,'' \emph{IEEE transactions on medical imaging}, vol.~23,
  no.~7, pp. 903--921, 2004.

\bibitem{raykar2010}
V.~C. Raykar, S.~Yu, L.~H. Zhao, G.~H. Valadez, C.~Florin, L.~Bogoni, and
  L.~Moy, ``Learning from crowds,'' \emph{Journal of Machine Learning
  Research}, vol.~11, no. Apr, pp. 1297--1322, 2010.

\bibitem{chen2019}
P.~Chen, B.~Liao, G.~Chen, and S.~Zhang, ``A meta approach to defend noisy
  labels by the manifold regularizer psdr,'' \emph{arXiv preprint
  arXiv:1906.05509}, 2019.

\bibitem{li2017b}
C.~Li, C.~Zhang, K.~Ding, G.~Li, J.~Cheng, and H.~Lu, ``Bundlenet: Learning
  with noisy label via sample correlations,'' \emph{IEEE Access}, vol.~6, pp.
  2367--2377, 2017.

\bibitem{bengio2009}
Y.~Bengio, J.~Louradour, R.~Collobert, and J.~Weston, ``Curriculum learning,''
  in \emph{Proceedings of the 26th annual international conference on machine
  learning}.\hskip 1em plus 0.5em minus 0.4em\relax ACM, 2009, pp. 41--48.

\bibitem{guo2018}
S.~Guo, W.~Huang, H.~Zhang, C.~Zhuang, D.~Dong, M.~R. Scott, and D.~Huang,
  ``Curriculumnet: Weakly supervised learning from large-scale web images,'' in
  \emph{Proceedings of the European Conference on Computer Vision (ECCV)},
  2018, pp. 135--150.

\bibitem{liu2017}
X.~Liu, S.~Li, M.~Kan, S.~Shan, and X.~Chen, ``Self-error-correcting
  convolutional neural network for learning with noisy labels,'' in \emph{2017
  12th IEEE International Conference on Automatic Face \& Gesture Recognition
  (FG 2017)}.\hskip 1em plus 0.5em minus 0.4em\relax IEEE, 2017, pp. 111--117.

\bibitem{li2017}
Y.~Li, J.~Yang, Y.~Song, L.~Cao, J.~Luo, and L.-J. Li, ``Learning from noisy
  labels with distillation,'' in \emph{Proceedings of the IEEE International
  Conference on Computer Vision}, 2017, pp. 1910--1918.

\bibitem{hinton2015}
G.~Hinton, O.~Vinyals, and J.~Dean, ``Distilling the knowledge in a neural
  network,'' \emph{arXiv preprint arXiv:1503.02531}, 2015.

\bibitem{malach2017}
E.~Malach and S.~Shalev-Shwartz, ``Decoupling" when to update" from" how to
  update",'' in \emph{Advances in Neural Information Processing Systems}, 2017,
  pp. 960--970.

\bibitem{yu2019}
X.~Yu, B.~Han, J.~Yao, G.~Niu, I.~Tsang, and M.~Sugiyama, ``How does
  disagreement help generalization against label corruption?'' in
  \emph{International Conference on Machine Learning}, 2019, pp. 7164--7173.

\bibitem{li2019}
J.~Li, Y.~Wong, Q.~Zhao, and M.~S. Kankanhalli, ``Learning to learn from noisy
  labeled data,'' in \emph{Proceedings of the IEEE Conference on Computer
  Vision and Pattern Recognition}, 2019, pp. 5051--5059.

\bibitem{tanaka2018}
D.~Tanaka, D.~Ikami, T.~Yamasaki, and K.~Aizawa, ``Joint optimization framework
  for learning with noisy labels,'' in \emph{Proceedings of the IEEE Conference
  on Computer Vision and Pattern Recognition}, 2018, pp. 5552--5560.

\bibitem{song2015}
S.~Song, K.~Chaudhuri, and A.~Sarwate, ``Learning from data with heterogeneous
  noise using sgd,'' in \emph{Artificial Intelligence and Statistics}, 2015,
  pp. 894--902.

\bibitem{rister2018}
B.~Rister, D.~Yi, K.~Shivakumar, T.~Nobashi, and D.~L. Rubin, ``Ct organ
  segmentation using gpu data augmentation, unsupervised labels and iou loss,''
  \emph{arXiv preprint arXiv:1811.11226}, 2018.

\bibitem{cciccek2016}
{\"O}.~{\c{C}}i{\c{c}}ek, A.~Abdulkadir, S.~S. Lienkamp, T.~Brox, and
  O.~Ronneberger, ``3d u-net: learning dense volumetric segmentation from
  sparse annotation,'' in \emph{International Conference on Medical Image
  Computing and Computer-Assisted Intervention}.\hskip 1em plus 0.5em minus
  0.4em\relax Springer, 2016, pp. 424--432.

\bibitem{ren2015}
S.~Ren, K.~He, R.~Girshick, and J.~Sun, ``Faster r-cnn: Towards real-time
  object detection with region proposal networks,'' in \emph{Advances in neural
  information processing systems}, 2015, pp. 91--99.

\bibitem{carass2017}
A.~Carass, S.~Roy, A.~Jog, J.~L. Cuzzocreo, E.~Magrath, A.~Gherman, J.~Button,
  J.~Nguyen, F.~Prados, C.~H. Sudre \emph{et~al.}, ``Longitudinal multiple
  sclerosis lesion segmentation: resource and challenge,'' \emph{NeuroImage},
  vol. 148, pp. 77--102, 2017.

\bibitem{commowick2018}
O.~Commowick, A.~Istace, M.~Kain, B.~Laurent, F.~Leray, M.~Simon, S.~C. Pop,
  P.~Girard, R.~Ameli, J.-C. Ferr{\'e} \emph{et~al.}, ``Objective evaluation of
  multiple sclerosis lesion segmentation using a data management and processing
  infrastructure,'' \emph{Scientific reports}, vol.~8, no.~1, pp. 1--17, 2018.

\bibitem{hashemi2019}
S.~R. Hashemi, S.~S.~M. Salehi, D.~Erdogmus, S.~P. Prabhu, S.~K. Warfield, and
  A.~Gholipour, ``Asymmetric loss functions and deep densely-connected networks
  for highly-imbalanced medical image segmentation: Application to multiple
  sclerosis lesion detection,'' \emph{IEEE Access}, vol.~7, pp. 1721--1735,
  2019.

\bibitem{allsbrook2001}
W.~C. Allsbrook~Jr, K.~A. Mangold, M.~H. Johnson, R.~B. Lane, C.~G. Lane, and
  J.~I. Epstein, ``Interobserver reproducibility of gleason grading of
  prostatic carcinoma: general pathologist,'' \emph{Human pathology}, vol.~32,
  no.~1, pp. 81--88, 2001.

\bibitem{arvaniti2018}
E.~Arvaniti, K.~S. Fricker, M.~Moret, N.~J. Rupp, T.~Hermanns, C.~Fankhauser,
  N.~Wey, P.~J. Wild, J.~H. Rueschoff, and M.~Claassen, ``Automated gleason
  grading of prostate cancer tissue microarrays via deep learning,''
  \emph{bioRxiv}, p. 280024, 2018.

\bibitem{karimi2019deep}
D.~Karimi, G.~Nir, L.~Fazli, P.~Black, L.~Goldenberg, and S.~Salcudean, ``Deep
  learning-based gleason grading of prostate cancer from histopathology
  images-role of multiscale decision aggregation and data augmentation.''
  \emph{IEEE journal of biomedical and health informatics}, 2019.

\bibitem{lampert2016}
T.~A. Lampert, A.~Stumpf, and P.~Gan{\c{c}}arski, ``An empirical study into
  annotator agreement, ground truth estimation, and algorithm evaluation,''
  \emph{IEEE Transactions on Image Processing}, vol.~25, no.~6, pp. 2557--2572,
  2016.

\bibitem{donovan2013}
T.~Donovan and D.~Litchfield, ``Looking for cancer: Expertise related
  differences in searching and decision making,'' \emph{Applied Cognitive
  Psychology}, vol.~27, no.~1, pp. 43--49, 2013.

\bibitem{nagpal2018}
K.~Nagpal, D.~Foote, Y.~Liu, E.~Wulczyn, F.~Tan, N.~Olson, J.~L. Smith,
  A.~Mohtashamian, J.~H. Wren, G.~S. Corrado \emph{et~al.}, ``Development and
  validation of a deep learning algorithm for improving gleason scoring of
  prostate cancer,'' \emph{arXiv preprint arXiv:1811.06497}, 2018.

\bibitem{robinson2016}
J.~W. Robinson, P.~C. Brennan, C.~Mello-Thoms, and S.~J. Lewis, ``The impact of
  radiology expertise upon the localization of subtle pulmonary lesions,'' in
  \emph{Medical Imaging 2016: Image Perception, Observer Performance, and
  Technology Assessment}, vol. 9787.\hskip 1em plus 0.5em minus 0.4em\relax
  International Society for Optics and Photonics, 2016, p. 97870K.

\bibitem{quekel1999}
L.~G. Quekel, A.~G. Kessels, R.~Goei, and J.~M. van Engelshoven, ``Miss rate of
  lung cancer on the chest radiograph in clinical practice,'' \emph{Chest},
  vol. 115, no.~3, pp. 720--724, 1999.

\bibitem{kundel1976}
H.~L. Kundel and G.~Revesz, ``Lesion conspicuity, structured noise, and film
  reader error,'' \emph{American Journal of Roentgenology}, vol. 126, no.~6,
  pp. 1233--1238, 1976.

\end{thebibliography}

\end{document}